\documentclass[pdflatex,sn-basic]{sn-jnl}

\usepackage{graphicx}
\usepackage{multirow}
\usepackage{longtable}
\usepackage{tabularx}
\usepackage{array} 
\usepackage{booktabs}
\usepackage[table]{xcolor}
\usepackage{amsmath,amssymb,amsfonts}
\usepackage{amsthm}
\usepackage{mathrsfs}
\usepackage[title]{appendix}
\usepackage{textcomp}
\usepackage{manyfoot}
\usepackage{algorithm}
\usepackage{algpseudocode}
\usepackage{algorithmicx}
\usepackage{listings}
\usepackage{natbib}
\usepackage{booktabs}
\usepackage{caption}
\usepackage{makecell}
\usepackage{rotating}
\usepackage{enumitem}
\usepackage{soul}    
\usepackage{pdfpages}
\usepackage{pifont}
\usepackage{comment}
\usepackage[T1]{fontenc}
\usepackage{upquote}
\usepackage{parskip}
\hypersetup{
    colorlinks=true,        
    linkcolor=blue,         
    citecolor=blue,         
    filecolor=blue,         
    urlcolor=red,           
}
\usepackage{hyperref}

\theoremstyle{thmstyleone}%
\theoremstyle{thmstyletwo}%

\theoremstyle{thmstylethree}%

\begin{document}

\title[Article Title]{A Survey of Classification Tasks and Approaches for Legal Contracts}

\author{\fnm{Amrita} \sur{Singh}}\email{amrita.singh1@unsw.edu.au}

\author{\fnm{Aditya} \sur{Joshi}}\email{aditya.joshi@unsw.edu.au}

\author{\fnm{Jiaojiao} \sur{Jiang}}\email{jiaojiao.jiang@unsw.edu.au}

\author{\fnm{Hye-young} \sur{Paik}}\email{h.paik@unsw.edu.au}

\affil{\orgdiv{School of Computer Science and Engineering}, \orgname{University of New South Wales (UNSW)}, \orgaddress{\street{Kensington}, \city{Sydney}, \postcode{2033}, \state{New South Wales}, \country{Australia}}}

\abstract{Given the large size and volumes of contracts and their underlying inherent complexity, manual reviews become inefficient and prone to errors, creating a clear need for automation. Automatic Legal Contract Classification (LCC) revolutionizes the way legal contracts are analyzed, offering substantial improvements in speed, accuracy, and accessibility. This survey delves into the challenges of automatic LCC and a detailed examination of key tasks, datasets, and methodologies. We identify seven classification tasks within LCC, and review fourteen datasets related to English-language contracts, including public, proprietary, and non-public sources. We also introduce a methodology taxonomy for LCC, categorized into Traditional Machine Learning, Deep Learning, and Transformer-based approaches. Additionally, the survey discusses evaluation techniques and highlights the best-performing results from the reviewed studies. By providing a thorough overview of current methods and their limitations, this survey suggests future research directions to improve the efficiency, accuracy, and scalability of LCC. As the first comprehensive survey on LCC, it aims to support legal NLP researchers and practitioners in improving legal processes, making legal information more accessible, and promoting a more informed and equitable society.}



\maketitle

\section{Introduction}\label{sec1}
Contracts are legally binding agreements that define the terms and conditions agreed upon by the involved parties \citep{fried2015contract}. Given the evolving legal regulations and the ever-increasing volume of legal documents within enterprises, both businesses and legal organizations are inundated with a large number of contracts, the scrutiny of which forms the basis for legal recommendations and organizational decision-making. As a result, automating various steps of the scrutiny process, such as the classification of legal contracts, emerges as a promising yet challenging endeavor. Legal Contract Classification (LCC) involves labeling different components of a contract, such as individual clauses, provisions (often referred to as sentences or paragraphs), or entire documents. These components are labeled based on tasks such as detecting risky clauses, recognizing ambiguity, or identifying the overall contract type (e.g., lease, consulting, software, consumer, or other contracts, each tailored for specific legal and business contexts). Traditionally, reviewing legal contractual documents is time-consuming and costly, a challenge that is particularly significant for individuals and organizations that cannot afford legal counsel. Automating legal contract classification reduces both time and costs, making legal reviews more efficient and accessible. This, in turn, helps address access-to-justice concerns by enabling individuals to avoid unfair terms without the need for expensive legal advice \citep{guha2024legalbench}.

Accurate classification of contracts is vital for numerous legal applications. It helps identify risky or unfair clauses \citep{lippi2019claudette, leivaditi2020benchmark, ruggeri2022detecting}, detect clauses with significant financial implications \citep{singh2024data}, and supports natural language inference tasks that uncover relationships between contract sections \citep{koreeda2021contractnli}. Furthermore, proper classification helps identify ambiguities in contract clauses \citep{singhal2024generating} and facilitates the tracking of responsibilities, deadlines, and actions tied to contract clauses, ensuring that stakeholders fulfill their obligations efficiently \citep{singh2024data}.

\begin{figure}[ht!]
\centering
\includegraphics[width=1\textwidth]{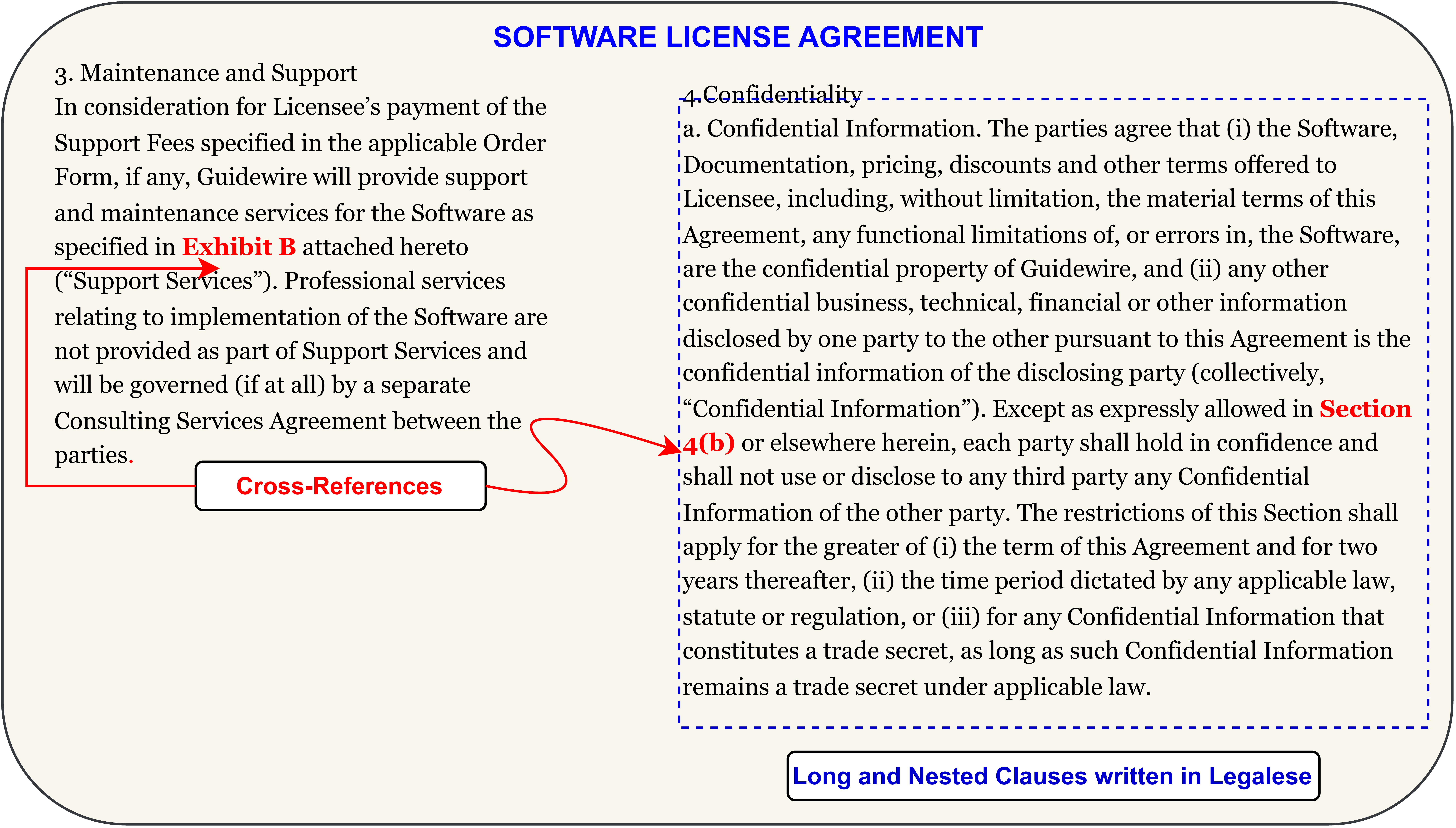}
\caption{Common Challenges in Legal Contract Clauses (\citeauthor{sec_license_agreement})}
\label{Fig0}
\end{figure}

Legal Contract classification plays a crucial role in improving governance, ensuring compliance, and enhancing operational efficiency at scale \citep{amoah2021effectiveness}. However, legal contract classification is more complex than standard text classification. Legal contracts are often written in complex, formal language with intricate legal terminology, known as "Legalese" \citep{katrak2022role}, with long and nested clauses, cross-references among clauses or documents, and complex contextual dependencies \citep{ariai2024natural}. These challenges are further compounded by jurisdictional variations, inconsistent formatting, and the long length of some contracts, which can span dozens or even hundreds of pages \citep{singh2024data}. Some of these challenges are illustrated in Figure \ref{Fig0}. With the increasing volume of contracts generated by IT outsourcing firms and businesses, sometimes reaching thousands each month, manual contract review becomes time-consuming and error-prone \citep{tauqeer2024infrastructure, khan2022challenges, singh2024data, singhal2024generating}. As a result, the demand for automated legal contract classification increases, offering a more efficient, accurate, and scalable solution to manage the growing workload.

Despite the growing significance of legal contract classification in the field of legal natural language processing, a notable gap remains in the literature regarding a comprehensive survey of the tasks involved in legal contract classification, the datasets available, and the methodologies employed. This survey aims to address this gap by providing an in-depth overview of the current state of legal contract classification, highlighting key challenges, methodologies, and potential future directions for this rapidly evolving field. By offering this comprehensive review, we aim to not only provide researchers with insights into the current state-of-the-art techniques but also offer valuable guidance to those new to the field of legal contract classification. To the best of our knowledge, this is the first survey focused exclusively on legal contract classification. The contributions of this paper are as follows:
\begin{enumerate}
    \item We provide a detailed overview of the various tasks involved in legal contract classification, identifying 7 classification tasks within this contractual domain.
    \item We review 14 legal contract classification datasets, organized according to the seven identified task categories. This includes 11 publicly available datasets, 1 non-publicly available dataset, and 2 proprietary datasets, all of which are crucial for future research. For each dataset, we summarize its key characteristics and present an overview in tabular form, ensuring easy access to relevant information for researchers. 
    \item We introduce a methodology-based taxonomy for legal contract classification tasks, organizing the approaches into three main categories: Traditional Machine Learning, Deep Learning, and Transformer-based approaches. We provide a broad overview as a figure and a more detailed analysis in tabular form for easy reference.
    \item We present the evaluation techniques used in legal contract classification and summarize the best-achieved performance results from previous works in a tabular format, offering a general overview of the performance in this area. 
    \item We discuss the primary challenges in legal contract classification and explore potential avenues for future advancements, providing a roadmap for continued progress in this field. 
\end{enumerate}
The rest of the paper is organized as follows: Section \ref{sec1.1} defines key terminology used throughout the paper. Section \ref{sec2} provides the background and motivation for conducting the survey on legal contract classification. Section \ref{sec3} outlines the review methodology and scope of the paper. Section \ref{sec4} discusses the identified legal contract classification tasks and datasets. Section \ref{sec5} focuses on the relevant approaches used for legal contract classification tasks. Throughout the paper, we include several summary tables that will be valuable for future research. Section \ref{sec6} presents the evaluation criteria employed to assess the performance of legal contract classification models and reports the results of the reviewed works. In Section \ref{sec8}, we address the primary challenges in legal contract classification and explore future research directions. Finally, Section \ref{sec9} concludes the survey. 

\section{Terminology and Definitions}\label{sec1.1}
This section defines key terms used throughout the paper, following common usage in prior research. These definitions help distinguish between closely related concepts in the legal NLP domain.

\textbf{\emph{Legal NLP Domain:}}
The legal NLP (Natural Language Processing) domain refers to the broader field of applying computational methods to legal texts \citep{ariai2024natural}. It includes various document types such as legislation, court rulings, case law, legal opinions, regulations, and contracts. Common tasks include legal judgment prediction, case outcome modeling, statute interpretation, and contract analysis. Contract analysis is one subfield within this broader domain.

\textbf{\emph{Contractual NLP Domain:}}
The contractual NLP domain is a specialized subarea within legal NLP that focuses exclusively on the analysis of contractual documents. Tasks in this domain include legal contract classification, summarization, question answering, and related applications. This narrower focus ensures that methods are tailored to the specific structure, semantics, and legal functions of contracts, while also accommodating variations across different jurisdictions.

\textbf{\emph{Contract Provision:}}
A contract provision refers to a paragraph or section within a contract that may contain one or more clauses. In prior work, the term \emph{provision} is often used interchangeably with \emph{paragraph} \citep{chalkidis2022lexglue, tuggener2020ledgar}.

\textbf{\emph{Contract Clause:}} A contract clause refers to a specific part of a contract, typically expressed at the sentence level. It outlines an obligation, condition, right, or requirement agreed upon by the parties \citep{indukuri2010mining}. In previous research, \emph{clause} are often referred to interchangeably as \emph{sentence} \citep{chalkidis2018obligation,  joshi2021domain, chalkidis2022lexglue, singhal2024generating, singh2024data}.

\textbf{\emph{Legal Contract Classification (LCC):}}
Legal Contract Classification (LCC) refers to the task of categorizing components of a contract into predefined classes. These components may include individual clauses, provisions (spanning sentences or paragraphs), or entire documents. LCC is foundational for automated contract review, risk analysis, and compliance checking.

\section{Background \& Motivation}\label{sec2}
LCC played an important role across industries by making the analysis and management of legal contracts easier. For legal professionals and compliance teams, it reduced the time needed to review contracts and helped ensure legal requirements were met efficiently. Contract managers and non-legal stakeholders relied on it to understand contract terms and manage responsibilities without constant legal support \citep{singhal2024generating}. Procurement teams used LCC to speed up vendor onboarding \citep{schuh2022profit}, while HR professionals applied it to review employment contracts, NDAs, and compliance terms \citep{armstead2015implementing}. Even individuals without legal training benefited from LCC, as it made contract analysis easier and supported informed decision-making without requiring deep legal knowledge.

In real-world use, LCC helped automatically identify important clauses, such as those related to termination or jurisdiction \citep{hendrycks2021cuad}. This improved the efficiency of tasks like due diligence, vendor onboarding, and partnership management. By reducing the need for manual review, LCC lowered the risk of missing critical details and improved compliance, especially in regulated industries like finance, healthcare, and data protection. When integrated into project workflows, LCC helped assign and track legal responsibilities, supporting better contract lifecycle management \citep{singh2024data}. This led to lower costs, fewer errors, and more reliable legal oversight, particularly in environments with limited legal resources.

To ground this survey in practical use, we examined how LCC had been applied in real-world scenarios involving contract review and analysis. AI-powered tools such as ROSS Intelligence \citep{ross2025} and Kira Systems \citep{kira2025} demonstrated LCC in action. These tools assisted in analyzing contracts and identifying important clauses, and they were adopted by various law firms and corporate legal departments to streamline contract review processes \citep{siino2025exploring}. Despite the growing importance of LCC, to the best of our knowledge, there was still no survey focused exclusively on legal contract classification.

Our review of existing surveys confirmed this gap.  
We conducted a review of existing surveys related to LCC. For example, \cite{chalkidis2019deep} examined the early adoption of deep learning in legal analytics, focusing on legal text classification, information extraction, and retrieval. Notably, only one study in this survey addressed the classification of contractual clauses to extract obligations and prohibitions. Similarly, \cite{villata2022thirty} reviewed research that applied machine learning and deep learning techniques in law. Here too, only one article focused on legal contract text classification, specifically using these techniques to identify unfair clauses in consumer contracts. From the surveys by \cite{chalkidis2019deep} and \cite{villata2022thirty}, we observed that while these works covered a broad range of legal texts including legislation, court cases, and contracts, the focus on legal contract classification and its associated tasks remained minimal. This trend is consistent across other studies as well, including those by \cite{ariai2024natural} and \cite{siino2025exploring}, where the emphasis is on the broader Legal NLP domain rather than on focused coverage of the Contractual NLP domain. As a result, the attention to LCC and its associated tasks remains limited. This analysis highlighted the need for more focused research in the area of legal contract classification.

\cite{montelongo2020tasks} conducted a bibliometric review of research articles on deep learning in the legal field from 1987 to 2020. They examined studies on nine tasks: classification, information extraction, information retrieval, summarization, text generation, feature extraction, preprocessing, and theoretical tasks, all involving various legal texts such as legislation, court cases, and contracts. They reported a 300\% increase in publications from 2016 to 2020 in the legal NLP field, with a particular focus on information extraction and classification, which together accounted for 39\% of the sample. Although they identified summarization and text generation as promising areas, they did not go into detail about the methodologies used by the studies in their review. Similarly, \cite{wang2024prompts} reviewed the use of large language models (LLMs) in contract drafting but did not discuss the methodologies and evaluation techniques in detail. While both reviews contributed to the field of Contractual NLP, their lack of in-depth assessments of methodologies and/or evaluation techniques limited a deeper understanding of the research. On the other hand, \cite{aejas2022review} reviewed the extraction of entities from legal texts, which is different from the classification task.

\cite{hassan2021addressing} reviewed research articles specifically addressing construction contracts and related tasks, such as entity extraction and classification. However, this narrow focus on construction-related legal contracts limited the generalizability of the findings, meaning that the insights gained may not apply to other types of contractual documents. Several other studies similarly focused on single types or domain-specific contracts and their associated tasks, including works by \cite{cardona2024bibliometric}, \cite{zhang2023comparing}, \cite{seo2022systematic}, \cite{zeberga2024digital}, and \cite{chung2023comparing}.

These surveys revealed several key gaps, which are summarized in Table \ref{Table1}. Below, we highlight the key limitations from Table \ref{Table1} that motivated us to conduct a comprehensive survey on legal contract classification:
\begin{enumerate}
\item \textbf{\emph{Lack of Focused Survey on Legal Contract Classification:}} Although there has been growing research in legal contract classification, no survey has specifically focused on this area. Most existing surveys focused on the broader field of Legal NLP and did not provide in-depth coverage of specific legal texts, such as contracts, and their related tasks. In fact, previous surveys gave limited attention to contractual text classification. As a result, methods, datasets, and tasks related to legal contract classification remained scattered across Contractual NLP domain, making it difficult to consolidate findings and track progress. This emphasized the need for a dedicated survey on legal contract classification.

\item \textbf{Under-explored Contemporary Paradigms:} Existing surveys that included legal contract classification as a small part of their review predominantly focused on traditional machine learning and deep learning methods. There was insufficient discussion of contemporary paradigms, such as LLM pre-training, prompting, and other techniques. As these technologies gained prominence, a more comprehensive exploration of their potential applications became essential.

\item \textbf{Domain-Specific Focus and Generalizability Issue:} Some existing reviews narrowed their scope to specific domains (e.g., construction contracts), which limited the applicability of their findings to other areas of legal contracts. A comprehensive survey would help generalize the findings across different contract types, enhancing the understanding of legal contract classification as a whole.
\end{enumerate}
\begin{table*}[ht!]
\centering
\LARGE
\caption{Overview of existing surveys including legal contractual text classification}
\label{Table1}
\resizebox{\textwidth}{!}{
\begin{tabular}{lllcccc}
\hline
\multicolumn{1}{c}{{\color[HTML]{FE0000} }} & \multicolumn{1}{c}{{\color[HTML]{FE0000} }} & \multicolumn{1}{c}{{\color[HTML]{FE0000} }} & \multicolumn{3}{c}{{\color[HTML]{FE0000} \textbf{Methodology}}} & {\color[HTML]{FE0000} } \\ \cline{4-6}
\multicolumn{1}{c}{\multirow{-2}{*}{{\color[HTML]{FE0000} \textbf{Survey}}}} & \multicolumn{1}{c}{\multirow{-2}{*}{{\color[HTML]{FE0000} \textbf{Focus}}}} & \multicolumn{1}{c}{\multirow{-2}{*}{{\color[HTML]{FE0000} \textbf{Task Type}}}} & \multicolumn{1}{l}{{\color[HTML]{3166FF} \textbf{\begin{tabular}[c]{@{}l@{}}Classical\\ Machine\\ Learning\end{tabular}}}} & \multicolumn{1}{l}{{\color[HTML]{3166FF} \textbf{\begin{tabular}[c]{@{}l@{}}Classical\\ Deep \\ Learning\end{tabular}}}} & \multicolumn{1}{l}{{\color[HTML]{3166FF} \textbf{\begin{tabular}[c]{@{}l@{}}Transformer\\based\end{tabular}}}} & \multirow{-2.7}{*}{{\color[HTML]{FE0000} \textbf{\begin{tabular}[c]{@{}c@{}}In-Depth Analysis of\\ Methods/Evaluation\\Techniques\\\end{tabular}}}} \\ \hline
\cite{chalkidis2019deep} & \begin{tabular}[c]{@{}l@{}}Legal NLP with \\ Limited Contract-\\ ual NLP coverage\end{tabular} & \begin{tabular}[c]{@{}l@{}}Various Tasks \\ (Including Limited \\ Study on LCC)\end{tabular} & {\color[HTML]{343434} \ding{51}} & {\color[HTML]{343434} \ding{51}} & {\color[HTML]{343434} \textcolor{red}{\ding{55}}} & {\color[HTML]{343434} \ding{51}} \\ \hline
\cite{montelongo2020tasks} & Legal NLP & \begin{tabular}[c]{@{}l@{}}Various Tasks \\ (Including LCC)\end{tabular} & {\color[HTML]{343434} \ding{51}} & {\color[HTML]{343434} \ding{51}} & {\color[HTML]{343434} \textcolor{red}{\ding{55}}} & {\color[HTML]{343434} \textcolor{red}{\ding{55}}} \\ \hline
\cite{hassan2021addressing} & \begin{tabular}[c]{@{}l@{}}Domain-Specific \\ Legal NLP \\ (Construction-related\\ legal texts)\end{tabular} & \begin{tabular}[c]{@{}l@{}}Various Tasks \\ (Including Limited \\ Study on LCC)\end{tabular} & {\color[HTML]{343434} \ding{51}} & {\color[HTML]{343434} \ding{51}} & {\color[HTML]{343434} \textcolor{red}{\ding{55}}} & {\color[HTML]{343434} \ding{51}} \\ \hline
\cite{aejas2022review} & Legal NLP & Entity Extraction & {\color[HTML]{343434} \ding{51}} & {\color[HTML]{343434} \ding{51}} & {\color[HTML]{343434} \textcolor{red}{\ding{55}}} & {\color[HTML]{343434} \ding{51}} \\ \hline
\cite{villata2022thirty} & \begin{tabular}[c]{@{}l@{}}Legal NLP with \\ Limited Contract-\\ ual NLP coverage\end{tabular} & \begin{tabular}[c]{@{}l@{}}Various Tasks \\ (Including Limited \\ Study on LCC)\end{tabular} & {\color[HTML]{343434} \ding{51}} & {\color[HTML]{343434} \ding{51}} & {\color[HTML]{343434} \textcolor{red}{\ding{55}}} & {\color[HTML]{343434} \ding{51}} \\ \hline
\cite{wang2024prompts} & Contractual NLP & Contract Drafting & {\color[HTML]{343434} \ding{51}} & {\color[HTML]{343434} \ding{51}} & {\color[HTML]{343434} \ding{51}} & {\color[HTML]{343434} \textcolor{red}{\ding{55}}} \\ \hline
{\color[HTML]{000000}\bfseries Our Survey} & 
{\color[HTML]{000000}\bfseries Contractual NLP} & 
{\color[HTML]{000000}\bfseries \makecell[l]{Legal Contract \\ Classification (LCC)}} & 
{\color[HTML]{343434}\bfseries \ding{51}} & 
{\color[HTML]{343434}\bfseries \ding{51}} & 
{\color[HTML]{343434}\bfseries \ding{51}} & 
{\color[HTML]{343434}\bfseries \ding{51}} \\ \hline
\end{tabular}
}
\end{table*}
Given these limitations, the need for a thorough, focused survey on legal contract classification became clear. To the best of our knowledge, this survey is the first to comprehensively address legal contract classification, aiming to fill these gaps by exploring the different tasks, datasets, methodologies, evaluation techniques, and challenges associated with this area. By doing so, we aim to advance the field and encourage future research, particularly in Contractual NLP and broadly in the Legal NLP domain.

\section{Review Methodology}\label{sec3}
\begin{figure}[ht!]
\centering
\includegraphics[width=1\textwidth]{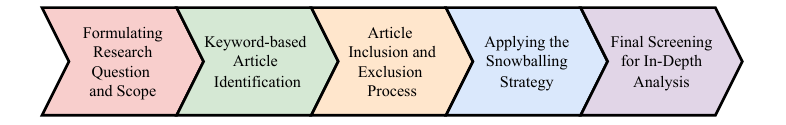}
\caption{Methodology of Review Process}\label{Fig1}
\end{figure}

This section outlines the review process and the identification of relevant studies. Figure \ref{Fig1} illustrates the steps of the review methodology.

The process begins by formulating the research question, which guides the scope of the study on legal contract classification. The research question is as follows:
\emph{What classification tasks, datasets, methods, models, evaluation metrics, and challenges shape legal contract classification, and how can future research be improved?}
To ensure a comprehensive review, a search strategy is employed to identify relevant studies addressing the research question. Search terms such as \emph{"Legal contract classification"}, \emph{"Legal clause extraction"}, and \emph{"Legal clause identification"}, along with their variations in spelling and tense, are used across several electronic literature databases, including \citeauthor{aclanthology_2025}, \citeauthor{ieee_2025}, \citeauthor{acm_portal_2025}, \citeauthor{springer_2025}, \citeauthor{google_scholar_2025}. This initial search results in a set of potentially relevant articles, which are then screened using inclusion and exclusion criteria, as detailed in Table \ref{Table2}. After applying these criteria, 52 articles remain. Studies that do not directly contribute to answering the research question are excluded, resulting in a set of 22 relevant articles. Although this may result in the omission of some studies, the goal is to capture a broad range of tasks and methodologies relevant to legal contract classification. In particular, we remove articles focused on legal analytics centered on contractual case prediction, case law analysis, or judicial decisions rather than legal contract classification; papers on legal contracts that do not involve any computational methods; studies on named entity recognition (NER) or information retrieval that do not include classification tasks; and works describing contract management systems that do not incorporate any computational techniques. To further ensure thoroughness, the snowballing approach, as outlined by \cite{wohlin2014guidelines}, is employed, which involves both backward snowballing (examining reference lists) and forward snowballing (checking citations) to identify additional relevant studies. After applying both backward and forward snowballing, a new set of 30 papers is identified.  The articles are then reviewed to confirm their alignment with the research question and inclusion criteria. Of these, 13 papers meet the criteria. Finally, the initial 22 relevant articles are combined with the 13 identified through snowballing, resulting in a total of 35 research articles selected for an in-depth critical review.

\begin{table*}[ht!]
\captionsetup{justification=centering}
\caption{Inclusion and Exclusion Criteria}
\centering
\label{Table2}
\resizebox{0.85\textwidth}{!}{
\begin{tabular}{ll}
\hline
\multicolumn{1}{c}{{\color[HTML]{3531FF} \textbf{Inclusion Criteria}}} & \multicolumn{1}{c}{{\color[HTML]{FE0000} \textbf{Exclusion Criteria}}} \\ \hline
\begin{tabular}[c]{@{}l@{}}CORE A*/A rated conferences\\ and Q1 journals research articles\\ \\ Workshop or Arxiv papers are\\ included only if they became\\ state-of-the-art for legal contract\\ classification or introduced a \\ novel method\\\\ Published between 1 January \\ 2010 to 31 October 2024\\ \\ Research articles that includes\\ English-language contracts \\ \\ Research articles published in\\ English\end{tabular} & \begin{tabular}[c]{@{}l@{}}Research articles published before \\ January 2010 and after November\\ 2024\\ \\ Research articles not published in\\ English\\ \\ Research articles that include\\ non-English language contracts\end{tabular} \\ \hline
\end{tabular}
}
\end{table*}
\section{Legal Contract Classification  Tasks and Datasets}\label{sec4}
\subsection{Tasks}
LCC is the process of organizing contract sentences (such as clauses or provisions) or entire documents into structured groups. Common LCC tasks include classifying the topic of a clause or provision \citep{tuggener2020ledgar}, identifying risky or unfair clauses \citep{lippi2019claudette, leivaditi2020benchmark, ruggeri2022detecting}, classifying deontic modality \citep{sancheti2022agent, chalkidis2018obligation, funaki2020contract}, identifying and classifying ambiguous clauses \citep{singhal2024generating}, and performing natural language inference \citep{koreeda2021contractnli}. This section introduces seven types of LCC tasks, provides examples, and explains the rationale behind the chosen labels, as detailed in Table \ref{Table-task}. 

\begin{table*}[ht!]
\caption{Tasks with Examples and Rationale}
\label{Table-task}
\resizebox{\textwidth}{!}{
\begin{tabular}{ccll}
\hline
{\color[HTML]{FE0000} \textbf{Task}} &
  {\color[HTML]{FE0000} \textbf{Classification}} &
  \multicolumn{1}{c}{{\color[HTML]{FE0000} \textbf{Example - Label}}} &
  \multicolumn{1}{c}{{\color[HTML]{FE0000} \textbf{Rationale}}} \\ \hline
 &
  \textbf{Multiclass} &
  \begin{tabular}[c]{@{}l@{}}Grantor agrees to pay the reasonable attorneys’ fees and legal \\ expenses incurred by Collateral Agent in the exercise of any \\ right or remedy available to it under this Agreement, whether \\ or not suit is commenced, including, without limitation, \\ attorneys’ fees and legal expenses incurred in connection with \\ any appeal of a lower court’s order or judgment - \color[HTML]{3531FF}\textbf{Expenses}\end{tabular} &
  \begin{tabular}[c]{@{}l@{}}The topic of contractual provision is labeled \\as \textit{\textbf{Expenses}} because it covers costs \\ related to the Collateral Agent's legal rights \\ including appeals, focusing on financial \\ obligations.\end{tabular} \\ \cline{2-4} 
\multirow{-10}{*}{\textbf{\begin{tabular}[c]{@{}c@{}}Topic\\ Classification\\\citep{tuggener2020ledgar}\\ \citep{chalkidis2022lexglue}\end{tabular}}} &
  \textbf{Multilabel} &
  \begin{tabular}[c]{@{}l@{}}The provisions of this Agreement, or any other Loan \\ Document, from time to time be amended, modified or waived, \\ if such amendment, modification or waiver is in writing and \\ consented to by the Borrower and both Lenders - \color[HTML]{3531FF}\textbf{Waivers}; \\ \color[HTML]{3531FF}\textbf{Amendments}\end{tabular} &
  \begin{tabular}[c]{@{}l@{}}The topic of contractual provision is labeled as\\ \textit{\textbf{Waiver}} and \textit{\textbf{Amendments}} because it \\ covers changes or waiver to the Agreement or \\ Loan Documents.\end{tabular} \\ \hline
 &
  \textbf{Multiclass} &
  \begin{tabular}[c]{@{}l@{}}1. This Contract is accepted to cancel through negotiation \\ together - \color[HTML]{3531FF}\textbf{Break option-Risky}\\ \\ \\ \\ 2. If either party is in breach of contract, it shall pay half \\ year rental as liquidated damage to the other party - \\ \color[HTML]{3531FF}\textbf{Damage-Risky}\\ \\ \\ 3. Lessee shall give Lessor not less than fifteen days a prior\\ written notice of the proposed assignment - \color[HTML]{3531FF}\textbf{Non-Risky}\end{tabular} &
  \begin{tabular}[c]{@{}l@{}}1. The clause is labeled as \textit{\textbf{Break option-Risky}} \\ because it allows sudden termination of the \\ contract through negotiation, causing an unfair \\ disadvantage for one party.\\ \\ 2. The clause is labeled as \textit{\textbf{Damage-Risky}} \\ because the fixed penalty may not match the \\ actual damages, leading to unfair compensation.\\ \\ \\ 3. The clause is labeled as \textit{\textbf{Non-Risky}} due to \\ reasonable notice period for assignment, allowing \\ time to respond.\end{tabular} \\ \cline{2-4} 
\multirow{-24}{*}{\textbf{\begin{tabular}[c]{@{}c@{}}Unfair/Risky\\ Clause\\ Identification\\ \citep{leivaditi2020benchmark}\\ \citep{lippi2019claudette}\\ \citep{ruggeri2022detecting}\end{tabular}}} &
  \textbf{Multilabel} &
  \begin{tabular}[c]{@{}l@{}}1. Any dispute, controversy or claim arising out of or  relating to\\ this EULA or the breach, termination or validity thereof shall be\\ finally settled at Rovio's discretion at your domicle's competent \\ courts; or by arbitration in accordance with the Rule for \\ Expedited Arbitration of the Arbitration Institute of the Finland \\ Chamber of Commerce. The arbitration shall be conducted in \\ Helsinki, Finland, in the English language - \\ \color[HTML]{3531FF}\textbf{Arbitration-Unfair; Jurisdiction-Fair}\\ \\ 2. Niantic further reserves the right to remove any User Content\\ from the Service at any time and without notice and for any \\ reason - \color[HTML]{3531FF}\textbf{Content Removal-Unfair}\\ \\ \\ 3. Outside the United States and Canada. If you acquired the \\ application in any other country, the laws of that country apply\\ - \color[HTML]{3531FF}\textbf{Choice of law-Fair}\end{tabular} &
  \begin{tabular}[c]{@{}l@{}}1. The clause is labeled as \textit{\textbf{Arbitration-Unfair}} \\ and \textit{\textbf{Jurisdiction-Fair}} because the arbitration is \\mandatory, restrictive, and controlled by the company, \\disadvantaging the consumer, while the jurisdiction \\allows the  consumer to resolve disputes in their local \\ courts, offering more fairness and accessibility.\\ \\ \\ 2. This clause is labeled as \textit{\textbf{Content}} \\ \textit{\textbf{Removal-Unfair}} because it gives the provider \\ full control to remove content at any time, for \\ any reason, and without notice.\\ \\ 3. The clause is labeled as \textit{\textbf{Choice of}} \\ \textit{\textbf{law-Fair}} because it applies the laws of the \\ consumer's country of the residence, ensuring \\ fairness in legal matters.\end{tabular} \\ \hline
 &
  \textbf{Multiclass} &
  \begin{tabular}[c]{@{}l@{}}1. The Supplier is obliged to meet and comply with the \\ Approved Requirements - \color[HTML]{3531FF}\textbf{Obligation}\\ \\ \\ 2. Nothing in this section will restrict either Party's right to \\ recruit - \color[HTML]{3531FF}\textbf{None}\\ \\ 3. Provider is not entitled to suspend this Agreement prior to \\ the lapse of the fifth year - \color[HTML]{3531FF}\textbf{Prohibition}\end{tabular} &
  \begin{tabular}[c]{@{}l@{}}1. The clause is labeled as \textit{\textbf{Obligation}} because it \\shows the supplier has a duty to meet and follow\\ the approved standards.\\ \\ 2. The clause is labeled as \textit{\textbf{None}} because it \\ does not impose any obligations or restrictions.\\ \\ 3. The clause is labeled as \textit{\textbf{Prohibition}} because it \\ stops the provider from suspending the \\ agreement before the fifth year.\end{tabular} \\ \cline{2-4} 
\multirow{-14}{*}{\textbf{\begin{tabular}[c]{@{}c@{}}Deontic \\ Modality\\ Classification \\\citep{chalkidis2018obligation}\\ \citep{sancheti2022agent}\end{tabular}}} &
  \textbf{Multilabel} &
  \begin{tabular}[c]{@{}l@{}}Tenant shall pay the rent to the Landlord and may use the \\ parking space - \color[HTML]{3531FF}\textbf{Obligation; Permission}\end{tabular} &
  \begin{tabular}[c]{@{}l@{}}The clause is labeled as \textit{\textbf{Obligation}} because the \\ tenant must pay rent, and \textit{\textbf{Permission}} because the \\ tenant is allowed, but not required, to use the \\ parking space.\end{tabular} \\ \hline
\textbf{\begin{tabular}[c]{@{}c@{}}Contractual \\ Ambiguity\\ Identification\\ \citep{singhal2024generating}\end{tabular}} &
  \textbf{Binary} &
  \begin{tabular}[c]{@{}l@{}}1. Snap will edit or write these articles as necessary to fit the \\ overall tone of the site - \color[HTML]{3531FF}\textbf{Ambiguous}\\ \\ \\ 2. Either Party may pledge this Agreement to Either Party may \\ pledge this Agreement to secure any credit facility or \\ indebtedness of such Party or its Affiliates without the consent \\ of the other Party - \color[HTML]{3531FF}\textbf{Not Ambiguous}\end{tabular} &
  \begin{tabular}[c]{@{}l@{}}1. The clause is labeled as \textit{\textbf{Ambiguous}} because it \\ lacks clear guidelines on what constitutes necessary \\ edits and how the overall tone of the site is defined.\\ \\ 2. The clause is labeled as \textit{\textbf{Not Ambiguous}} because \\ it clearly state the either party can pledge the \\ agreement as collateral without needing the other \\ party's consent, leaving no room for confusion.\end{tabular} \\ \hline
\textbf{\begin{tabular}[c]{@{}c@{}}Norm\\ Conflict\\ Identification\\ \citep{aires2018norm}\end{tabular}} &
  \textbf{Binary} &
  \begin{tabular}[c]{@{}l@{}}1. Notwithstanding the foregoing, Ligand shall remain \\ responsible for the Firm Commitment portion of the Rolling \\ Forecast. - \color[HTML]{3531FF}\textbf{Conflicting Norm}\\ \\ 2. Notwithstanding the foregoing, Ligand shall not remain \\ responsible for the Firm Commitment portion of the Rolling\\ Forecast - \color[HTML]{3531FF}\textbf{Conflicting Norm}\end{tabular} &
  \begin{tabular}[c]{@{}l@{}}These two norms are labeled as \textit{\textbf{Conflicting Norm}} \\ because the first one makes Ligand commit to \\ orders in advance, while the second one says \\ Ligand is not responsible for the forecasted \\ quantities, which creates confusion about their \\ actual responsibilities.\end{tabular} \\ \hline
\textbf{\begin{tabular}[c]{@{}c@{}}Obligatory\\ Clause\\ Classification\\ \citep{singh2024data}\end{tabular}} &
  \textbf{Multilabel} &
  \begin{tabular}[c]{@{}l@{}}The vendor must ensure that communications and server \\ rooms are secured with an access card system - \\ \color[HTML]{3531FF}\textbf{Information security-Physical security-Work area restriction}\end{tabular} &
  \begin{tabular}[c]{@{}l@{}}The clause is classified into \textit{\textbf{Information Security}}\\ (protecting sensitive information access), \textit{\textbf{Physical}}\\ \textit{\textbf{Security}} (controlling access to space), \textit{\textbf{Work area}}\\ \textit{\textbf{restriction}} (ensuring authorized access to critical\\ infrastructure).\end{tabular} \\ \hline
  \textbf{\begin{tabular}[c]{@{}c@{}}Natural Language \\ Inference for\\ Contracts\\ \citep{koreeda2021contractnli}\end{tabular}} &
  \textbf{Multiclass} &
  \begin{tabular}[c]{@{}l@{}}\textbf{Context:} Confidential Information: means all confidential\\ information (however recorded, preserved or disclosed)\\ disclosed by a Party or its Representatives to the other Party \\ and that Party's Representatives including but not limited to:\\
(a) the fact that discussions and negotiations are taking place \\concerning the Purpose and the status of those discussions and \\negotiations;\\
(b) the existence and terms of this Agreement; \\ \textbf{Hypotheses:} Receiving Party shall not disclose the fact that \\ Agreement was agreed or
negotiated -\color[HTML]{3531FF}\textbf{Entailment} \end{tabular} &
  \begin{tabular}[c]{@{}l@{}}The hypothesis is classified into \textit{\textbf{Entailment}} because\\ it directly follows the context, such as the confidentiality \\clause prohibiting disclosure of negotiations or \\agreement details.\end{tabular} \\ \hline
  
\end{tabular}
}
\end{table*}

\emph{\textbf{Topic Classification:}} The task aims to identify the principal theme or topics within contract clauses, provisions, or documents. This task can be either a multi-class \citep{chalkidis2022lexglue} or multi-label \citep{tuggener2020ledgar} classification problem, depending on the formulation of the problem.

\emph{\textbf{Risky/Unfair Clause Identification:}} This task focuses on identifying contract clauses that pose risks or are unfair to one or more parties involved in the agreement. It can be either a multi-class \citep{leivaditi2020benchmark} or multi-label \citep{lippi2019claudette, ruggeri2022detecting} classification problem.

\emph{\textbf{Deontic Modality Classification:}} This task involves classifying contract clauses into deontic categories, such as obligations, permissions, prohibitions, or other related categories. These clauses are typically expressed using modal verbs (e.g., must, should, may, cannot), which indicate what is required, allowed, or forbidden in the contract. The task can be approached as either a multi-class \citep{chalkidis2018obligation} or a multi-label \citep{sancheti2022agent} classification problem.

\emph{\textbf{Contractual Ambiguity Identification:}} The task involves identifying contract clauses that contain ambiguous language and classifying these ambiguous clauses into types, such as vagueness, incompleteness, referential, semantic, syntactic, lexical, or other forms \citep{massey2014identifying}, to determine the source of the ambiguity. It can be approached as either a binary (ambiguous/unambiguous) \citep{singhal2024generating}, multi-class, or multi-label classification problem.

\emph{\textbf{Norm Conflict Identification:}} Contracts use deontic statements (norms) to define terms and conditions, and conflicting norms can invalidate the contract. The task identifies contradictions between norms in a contract, such as conflicting obligations, permissions, or prohibitions. It involves identifying inconsistencies between clauses, such as when one clause requires an action while another forbids it, to ensure the contract is clear and logically consistent. Two norms can conflict if they have different deontic terms (such as obligation, permission, or prohibition) but involve the same action \citep{aires2017norm}. This task is typically framed as a binary classification problem (conflict/no conflict) \citep{aires2018norm} or a multi-class classification problem (e.g., obligation vs. prohibition, obligation vs. permission, permission vs. prohibition) \citep{aires2017norm, aires2021norm}.

\emph{\textbf{Obligatory Clause Classification:}} The task involves classifying obligatory clauses in contract documents based on their function, such as IT-specific requirements (e.g., security or privacy), governance-related requirements, or architectural mandates crucial for project success. This task can be approached as a multi-label \citep{sainani2020extracting, singh2024data} classification problem to identify and categorize different types of obligations.

\emph{\textbf{Natural Language Inference (NLI) for Contracts:}} This task involves determining whether a hypothesis (e.g., "Some obligations may survive termination") is supported by, contradicts, or is neutral to a contract. The system also identifies specific parts of the contract that support the decision \citep{koreeda2021contractnli}.

\subsection{Datasets}\label{sec5.2}
\begin{figure}[ht!]
\centering
\includegraphics[width=1\textwidth]{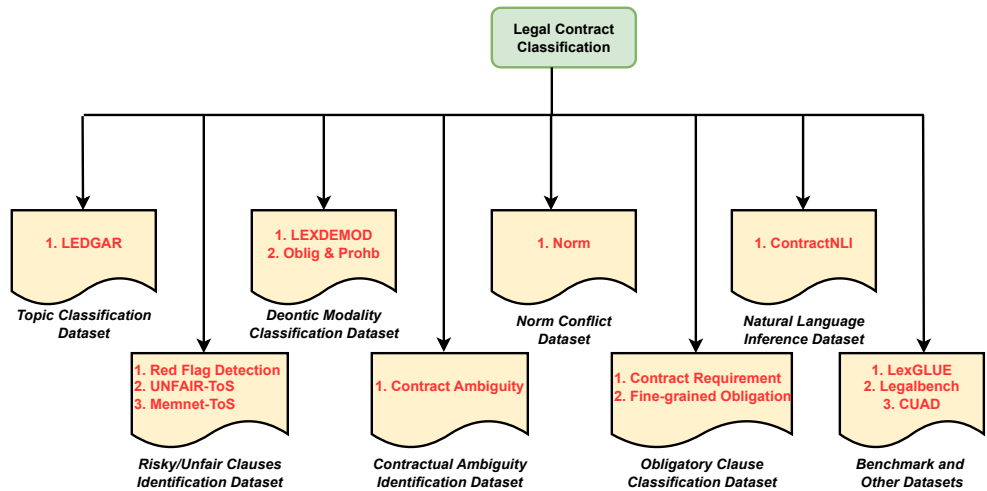}
\caption{Overview of Datasets Grouped According to Legal Contract Classification Tasks}\label{Fig2}
\end{figure}

This section provides an overview of commonly used datasets in LCC research. The availability of labeled datasets is a key factor driving rapid progress in this field. The datasets are organized according to the task categories outlined in Figure \ref{Fig2}. 

For each dataset, we summarize key characteristics and provide an overview in Table \ref{Table3}, including the source from which contract data are extracted, the types of contracts included in the dataset, the country of origin, the annotation schemes, dataset size, number of categories, and the available access links along with the dataset name.

\subsubsection{Topic Classification Dataset}
\emph{\textbf{LEDGAR:}} The LEDGAR \citep{tuggener2020ledgar} dataset is a multi-label corpus designed for the analysis and classification of legal contract provisions, with a particular focus on contracts filed with the U.S. Securities and Exchange Commission (SEC) through its EDGAR system. The SEC's Electronic Data Gathering, Analysis, and Retrieval (EDGAR) system provides access to financial documents, including annual reports, registration statements, and other filings made by publicly traded companies and various organizations. These documents often contain detailed contracts, which serve as a rich source for legal text analysis. The LEDGAR corpus is primarily built around Exhibit-10 material contracts, which are a common type of agreement found in SEC filings. These contracts include key legal documents such as shareholder agreements, employment contracts, non-disclosure agreements, and so on. The advantage of focusing on these agreements is that they frequently contain provisions of a similar nature (e.g., governing law, dispute resolution, confidentiality clauses), which appear across a wide range of contracts. A total of 60,540 Exhibit-10 contracts, filed between 2016 and 2019, are selected, resulting in 846,274 provisions. These provisions are semi-automatically annotated with their principal topics, producing 12,608 distinct labels through a combination of automated and heuristic methods.

\subsubsection{Risky/Unfair Clauses Identification Datasets}
\emph{\textbf{Red Flag Detection:}} The Red Flag Detection \citep{leivaditi2020benchmark} dataset is a multi-class corpus designed to identify and classify potential red flags in contract clauses that may pose risks to one or more parties involved. This dataset is created using lease agreements publicly available through the U.S. SEC's EDGAR database. It consists of 53,232 clauses extracted from 179 lease agreements. Of these, 51,990 clauses are manually annotated as the negative class (neutral/non-risky clauses), and 1,242 are annotated as the positive class (red flag/risky clauses) in consultation with legal professionals specializing in real estate law. The red flag/risky clauses are further classified into 19 types, such as \emph{sublease, right of first refusal to purchase (ROFR to purchase), right of first refusal to lease (ROFR to lease), as-is reinstatement, option to purchase, no obligations to operate, bank guarantee, rent review, non-transferable security, warranties, compulsory reconstruction, C.V., change of control, break option, termination, indexation, landlord repairs, damage, and expansion}.

\emph{\textbf{UNFAIR-ToS:}} The UNFAIR-ToS dataset \citep{lippi2019claudette} is a multi-label corpus designed to identify unfair clauses in the terms of service (ToS) of online platforms such as Facebook, Fitbit, Google, Instagram, LinkedIn, and others. It focuses on detecting potentially unfair clauses that create an imbalance in the rights and obligations between parties, typically disadvantaging the consumer, in accordance with the definitions outlined in EU Directive 93/13 on Unfair Terms in Consumer Contracts \citep{reich2014european}. The dataset consists of 50 ToS documents, totaling 12,011 clauses, with 8.6\% (1,032 clauses) flagged as potentially unfair. These clauses are manually categorized into one or more of the following eight types: \emph{arbitration, unilateral change, content removal, jurisdiction, choice of law, limitation of liability, unilateral termination, and contract by using}. Each clause type is assigned a fairness score of 1 (fair), 2 (potentially unfair), or 3 (unfair), resulting in eight types of unfair clauses.

\emph{\textbf{Memnet-ToS:}} The Memnet-ToS \citep{ruggeri2022detecting} dataset is a multi-label corpus designed to identify and analyze potentially unfair clauses in the terms of service (ToS) of online platforms. It includes 100 ToS documents and contains 21,063 clauses, of which 11.1\% (2,346 clauses) are flagged as potentially or clearly unfair. Over a period of eighteen months, four legal experts manually tag the potentially unfair clauses according to the guidelines outlined by \cite{lippi2019claudette} and categorize the clauses into one or more of the following five types: \emph{limitation of liability}, \emph{content removal}, \emph{unilateral termination}, \emph{unilateral changes}, and \emph{arbitration}. Each clause is assigned a score of 0 (fair) or 1 (unfair), resulting in five types of unfair clauses.

\subsubsection{Deontic Modality Classification Datasets}
\emph{\textbf{LEXDEMOD:}} The LEXDEMOD \citep{sancheti2022agent} dataset is a multi-label corpus designed to classify contract clauses into deontic categories. These clauses are typically expressed with modal verbs (e.g., must, should, may, cannot), indicating what is required, allowed, or forbidden. The dataset uses contracts from the LEDGAR dataset \citep{tuggener2020ledgar}, which includes various contract types (e.g., shareholder agreements, employment contracts, leases, and non-disclosure agreements) sourced from the EDGAR system. LEXDEMOD contains 7,092 clauses from 23 lease contracts and 8,230 span annotations. Each clause is manually annotated with one or more of seven types: \emph{obligation, entitlement, prohibition, permission, no obligation, no entitlement, and none}. These annotations specify the modality type as it applies to a particular contracting party or agent, along with the corresponding modal triggers.

\emph{\textbf{Oblig \& Prohb:}} The Oblig \& Prohb \citep{chalkidis2018obligation} dataset is a multi-class corpus consisting of 100 randomly selected English service agreements. It includes 45,144 clauses extracted from these agreements, manually annotated with six gold classes: \emph{none, obligation, prohibition, obligation list intro, obligation list item, and prohibition list item}. The annotation is carried out by five law students and reviewed by a paralegal expert, following strict guidelines.

\subsubsection{Contractual Ambiguity Identification Dataset}
\emph{\textbf{Contract Ambiguity:}} The Contract Ambiguity \citep{singhal2024generating} dataset is a binary classification corpus aimed at identifying ambiguous contract clauses. It consists of 1,000 clauses, which are sourced from the CUAD dataset \citep{hendrycks2021cuad}. These clauses are annotated by four non-legal stakeholders, each with over five years of experience in contracts. Each clause is labeled as either \emph{ambiguous} or \emph{unambiguous}, with 524 clauses classified as ambiguous and 476 as unambiguous.

\subsubsection{Norm Conflict Identification Datasets}
\emph{\textbf{Norm:}} The Norm \citep{aires2017norm, aires2018norm, aires2021norm} dataset consists of 1,193 manually annotated contract clauses, with 699 labeled as norms and 494 as non-norms. It also includes a semi-automatically constructed corpus containing 111 norm pairs with conflicting norms. Two volunteers contribute to creating these conflicts: the first volunteer inserts 94 conflicts across 10 contracts by changing modal verbs (e.g., altering "must" to "may"), resulting in 13 conflicts between permission and prohibition, 36 between permission and obligation, and 46 between obligation and prohibition. The second volunteer introduces 17 conflicts across 6 contracts by modifying deontic actions, leading to 4 conflicts between permission and prohibition, 8 between permission and obligation, and 5 between obligation and prohibition.

\subsubsection{Obligatory Clause Classification Datasets}
\emph{\textbf{Contract Requirement:}} The Contract Requirement \citep{sainani2020extracting} dataset is a multi-label corpus designed to extract and classify requirements from obligation clauses in software engineering contracts. Contracts typically contain two main types of clauses: obligations and non-obligations. Obligations are mandatory clauses that express actionable requirements, which can be IT-specific (e.g., security or privacy), governance-related, or architecturally significant. These obligations are essential for the successful delivery of a project. In contrast, non-obligations include information, definitions, or factual statements that do not translate into actionable requirements. The dataset consists of 20 expired contracts from 9 application domains, including healthcare, automotive, and finance. It contains a total of 18,614 clauses, of which 5,472 are obligation clauses. These obligation clauses are further categorized into 14 requirement types: \emph{project delivery, information security, legal process, screening/onboarding, data privacy, vendor corporate, improvement and innovation, personnel allocation, HR client policy, HR laws, third-party IP licensing, vendor IP licensing, export laws, and standards}. The remaining 13,142 clauses are non-obligation clauses.

\emph{\textbf{Fine-grained Obligation:}} The Fine-grained Obligation dataset \citep{singh2024data} is a multi-label corpus designed to extract and classify obligations from software engineering contracts. It contains 50 contracts from 13 sectors (e.g., healthcare, automotive, finance, telecom), totaling 57,200 statements, including 16,538 obligation clauses and 40,662 non-obligation clauses. Obligation clauses are annotated using a fine-grained structure called the Business Function-Responsibility-Customer Need (BF-R-CN) triplet, with 152 distinct triplets. The Business Function refers to the department responsible for fulfilling the obligation (e.g., Security, Legal), Responsibility is the duty to fulfill the obligation (e.g., Compliance, Audit), and Customer Need represents the specific requirement (e.g., Price Review for Audit). For instance, the obligation \emph{"The vendor shall comply with all security requirements"} is annotated as \textbf{Security} (BF) - \textbf{Compliance} (R) - \textbf{Customer\_specific\_policy\_adherence} (CN).

\subsubsection{NLI Dataset}
\emph{\textbf{ContractNLI:}} The ContractNLI \citep{koreeda2021contractnli} dataset is designed for document-level Natural Language Inference (NLI) to automate contract review, specifically for non-disclosure agreements (NDAs) crawled from the US SEC's EDGAR system. It contains 607 annotated contracts and 17 hypotheses. The task involves classifying whether each hypothesis is entailed by, contradicts, or is not mentioned (neutral to) the contract, along with identifying evidence spans that support the classification. 

\subsubsection{Benchmark and Other Datasets}
\emph{\textbf{LexGLUE:}} The LexGLUE \citep{chalkidis2022lexglue} dataset includes seven datasets for evaluating legal Natural Language Understanding (NLU) tasks, with two key datasets focused on contract classification: LEDGAR \citep{tuggener2020ledgar} and UNFAIR-ToS \citep{lippi2019claudette}. For LexGLUE, the LEDGAR dataset is simplified to include 80,000 contract provisions from SEC filings, categorized into the 100 most frequent themes, while UNFAIR-ToS focuses on identifying unfair terms in 50 Terms of Service documents, annotated with 8 types of unfair clauses. Both datasets are split chronologically into training, development, and test sets, enabling focused evaluation of models for legal contract classification and the identification of unfair contractual terms. When referring to LexGLUE in this paper, we specifically discuss these two legal contract classification datasets, as the other datasets in LexGLUE pertain to different domains or tasks, which fall outside the scope of our survey.

\begin{table*}[ht!]
\caption{Overview of the Legal Contract Classification Datasets}
\centering
\label{Table3}
{\footnotesize
\setlength{\arraycolsep}{-3pt}
\renewcommand{\arraystretch}{1.2} 
\setlength{\tabcolsep}{3pt} 
\begin{tabularx}{\textwidth}{| >{\centering\arraybackslash}p{1.3cm} | >{\centering\arraybackslash}p{1.6cm} | >
{\centering\arraybackslash}p{1.55cm} | >{\centering\arraybackslash}p{1.5cm} | >
{\centering\arraybackslash}p{1.1cm} | >
{\centering\arraybackslash}p{1.55cm} | >{\centering\arraybackslash}X | >
{\centering\arraybackslash}p{1cm} | }
\hline
\textbf{Author, \newline Year} & \textbf{\centering Dataset Name}  &  {\textbf{\multirow{1}{=}[-1ex]{\centering Source}}} & \textbf{\multirow{1}{=}[-1ex]{\centering Type}} & \textbf{\multirow{1}{=}[-1ex]{Country}} & \textbf{\multirow{1}{=}[-1ex]{Annotation}}  & \textbf{\multirow{1}{=}[-1ex]{\centering Size}} & \textbf{\multirow{1}{=}[-1ex]{\multirow{1}{=}[-0.5ex]{\centering Classes}}}
 \\\hline
\multirow{1}{=}[-1.5ex]{\centering \cite{tuggener2020ledgar}} & {\multirow{1}{=}[-4ex]{\makecell{\href{https://drive.switch.ch/index.php/s/j9S0GRMAbGZKa1A}{LEDGAR}}}} &  & \multirow{1}{=}[-1.5ex]{\centering Exhibit-10 material contract}  &  & \multirow{1}{=}[-3ex]{\centering Semi-Automatic} & {\centering 60,540 contracts (846,274 paragraphs)} & \multirow{1}{=}[-4ex]{\centering 12,608}\\
\cline{1-2}\cline{4-4}\cline{6-8}

\centering{\cite{leivaditi2020benchmark}} &	{\multirow{1}{=}[-1.5ex]{\makecell[cc]{\href{https://uvaauas.figshare.com/articles/dataset/ALeaseBert/19732993/1}{Red Flag} \\ \href{https://uvaauas.figshare.com/articles/dataset/ALeaseBert/19732993/1}{Detection}}}}	& &	  \multirow{1}{=}[-5ex]{\centering Lease contract} &	 &		& {\centering 179 contracts (53,232  sentences)} & \multirow{1}{=}[-3ex]{\centering 19} \\				
\cline{1-2}\cline{7-8}

\cite{sancheti2022agent} & {\multirow{1}{=}[-0.3ex]{\makecell[cc]{\href{https://github.com/adobe-research/LexDeMod/tree/main/deontic_data}{LEXDEMOD}}}} &	\multirow{1}{=}[0.1ex]{\centering SEC's EDGAR Database} & & \multirow{1}{=}[-1.5ex]{\centering US} & &	{\centering 23 contracts (7,092  sentences)} &	\multirow{1}{=}[-3ex]{\centering 7}\\			
\cline{1-2}\cline{4-4}\cline{7-8}

\cite{singhal2024generating} & {{\multirow{1}{=}[-1.5ex]{\centering \href{https://github.com/anmolsinghal98/Clarification-Question-Generation-for-Contracts}{Contract Ambiguity}}}} 	&  & \multirow{1}{=}[-1.5ex]{\centering Multiple contract (Affiliate, Consulting, etc.)}& & &	{\centering 25 contracts (1000 sentences)} &	\multirow{1}{=}[-3ex]{\centering 2}\\
\cline{1-2}\cline{7-8}

{\centering \cite{hendrycks2021cuad}} & {\makecell[cc]{\multirow{1}{=}[-0.1ex]{\;\;\;\;\;\href{https://www.atticusprojectai.org/cuad}{CUAD}}}} & &  & &  & {\centering 510 contracts (13,101 sentences)} & \multirow{1}{=}[-2.5ex]{\centering 41}\\
\cline{1-5}\cline{7-8}

\multirow{1}{=}[-3ex]{\centering \cite{lippi2019claudette}} & {\multirow{1}{=}[-3ex]{\centering\href{http://claudette.eui.eu/ToS.zip}{UNFAIR-ToS}}} & \multirow{1}{=}[-8ex]{\centering Online  Platforms} & \multirow{1}{=}[-5ex]{\centering Terms of Service  
(Consumer Contracts)} & & \multirow{1}{=}[-13ex]{\centering Manual} & {\centering 50 ToS contracts (12,011 sentences)} & \multirow{1}{=}[-4ex]{\centering 8}\\
\cline{1-2}\cline{7-8}

\multirow{1}{=}[-1.5ex]{\centering \cite{ruggeri2022detecting}} & {\makecell[bc]{\multirow{1}{=}[-1ex]{\centering \href{https://github.com/federicoruggeri/Memnet_ToS}{Memnet-ToS}}}}  &  &  & \multirow{1}{=}[-2ex]{\centering EU} &  & {\centering 100 ToS contracts (21,063 sentences)} & \multirow{1}{=}[-4ex]{\centering 5}\\
\cline{1-2}\cline{3-4}\cline{7-8}


\cite{chalkidis2018obligation} & \multirow{1}{=}[-1.5ex]{\centering \textcolor{red}{Oblig \& Prohb}} & \multirow{1}{=}[-1.5ex]{\centering Not Specified} & \multirow{1}{=}[-1.5ex]{\centering Service agrements} &  &  & {\centering 100 contracts (45,144 sentences)} & \multirow{1}{=}[-2.5ex]{\centering 6}\\
\cline{1-2}\cline{3-5}\cline{7-8}

\multirow{1}{=}[-1.5ex]{\centering \cite{aires2017norm}} & {\multirow{1}{=}[-2.5ex]{\centering \href{https://github.com/JoaoPauloAires/norm-dataset/tree/master}{Norm}}} & \multirow{1}{=}[-1.5ex]{\centering Onecle Database} & {\centering Multiple (business, lease, etc.)} & \multirow{1}{=}[-2.5ex]{\centering Australia} &  & \multirow{1}{=}[-1.5ex]{\centering 1193 and 111 sentences} & \multirow{1}{=}[-2ex]{\centering 2}\\
\cline{1-2}\cline{3-5}\cline{7-8}

{\multirow{1}{=}[-1.5ex]{\centering \cite{koreeda2021contractnli}}} & {\multirow{1}{=}[-5ex]{\centering \href{https://stanfordnlp.github.io/contract-nli/}{ContractNLI}}}  & {\centering SeC's EDGAR Database, 
Internet Search} & \multirow{1}{=}[-2.5ex]{\centering Non-disclosure agreements} & \multirow{1}{=}[-3.5ex]{\centering US, Others} &  & \multirow{1}{=}[-4ex]{\centering 607\;\; contracts} & \multirow{1}{=}[-5ex]{\centering 3}\\
\cline{1-2}\cline{3-5}\cline{7-8}

\cite{sainani2020extracting} & \multirow{1}{=}[-1.5ex]{\centering \textcolor{red}{Contract Requirement}}
 & \multirow{1}{=}[-6ex]{\centering Organization Database} & \multirow{1}{=}[-4ex]{\centering Software Engineering 
contracts} & \multirow{1}{=}[-6ex]{\centering Multiple} &  & {\centering 20 contracts (18,614 sentences)} & \multirow{1}{=}[-2.5ex]{\centering 14}\\
\cline{1-2}\cline{7-8}

\multirow{1}{=}[-2ex]{\centering \cite{singh2024data}} & \multirow{1}{=}[-1.5ex]{\centering \textcolor{red}{Fine-grained Obligation}}  
 &   &  &  &  & {\centering 50 contracts 
(16,538 sentences)} & \multirow{1}{=}[-2ex]{\centering 152}\\
\cline{1-8}

\cite{chalkidis2022lexglue} & {\multirow{1}{=}[-2.5ex]{\centering \href{https://huggingface.co/datasets/lex_glue}{LexGLUE}}} & \multirow{1}{=}[-5ex]{\centering Multiple} & \multirow{1}{=}[-5ex]{\centering Multiple} & \multirow{1}{=}[-5ex]{\centering US, EU} & \multirow{1}{=}[-4ex]{\centering Not Applicable} & \multirow{1}{=}[-2ex]{\centering Compilation of Different Contract Datasets} & \multirow{1}{=}[-2ex]{\centering 2 tasks: [9,100]}\\
\cline{1-2}\cline{8-8}

\cite{guha2024legalbench} & {\multirow{1}{=}[-1ex]{\centering\href{https://huggingface.co/datasets/nguha/legalbench}{\tiny LEGALBENCH}}}  &  &  &  &  &  & \multirow{1}{=}[-0.55ex]{41 tasks : [2-8]}\\
\hline
\end{tabularx}
}
\end{table*}

\emph{\textbf{LEGALBENCH:}} The LEGALBENCH \citep{guha2024legalbench} dataset aims to establish an open and collaborative legal reasoning benchmark for the few-shot evaluation of LLMs. It represents the first step toward constructing an interdisciplinary, collaborative legal reasoning benchmark for the English language and evaluates 20 LLMs across 162 legal tasks from 36 different data sources. These 162 tasks vary in sample size: 125 tasks have between 50 and 500 samples, 29 tasks have between 500 and 2,000 samples, and only 8 tasks have more than 2,000 samples. For contract classification-related tasks, the dataset includes lightweight tasks, such as the Contract QA, CUAD, J.Crew blocker, Unfair Terms of Service, and Contract NLI tasks, which have smaller sample sizes compared to the original tasks in the CUAD \citep{hendrycks2021cuad}, Unfair Terms of Service \citep{lippi2019claudette}, and Contract NLI \citep{koreeda2021contractnli} datasets. When referring to LEGALBENCH in this paper, we specifically discuss the legal contract classification datasets above, as the other datasets pertain to different domains or tasks, which fall outside the scope of our survey.

\emph{\textbf{CUAD:}} The Contract Understanding Atticus Dataset (CUAD) \citep{hendrycks2021cuad} is designed to support Legal NLP research by automating clause identification and classification. CUAD is a legal corpus containing 13,101 labeled clauses from 510 commercial contracts, sourced from the EDGAR system. It covers 25 contract types (e.g., Affiliate, Consulting, Franchise, Licensing) and 41 legal clause categories (e.g., IP Ownership, Non-Compete, Warranty Duration, Termination for Convenience). The dataset includes a CUAD\_v1 file, a SQuAD-2.0 style \citep{rajpurkar-etal-2018-know} JSON for question-answering, and 28 Excel files for specific clause categories.

\section{Legal Contract Classification Methods}\label{sec5}
This section presents an in-depth analysis of the methodology used in legal contract classification tasks, as outlined in Table \ref{Task-Method-Mapping}. The table reveals that legal contract classification primarily focuses on three key tasks: topic classification, unfair/risky clause classification, and deontic modality classification. Between 2010 and 2019, classical machine learning and deep learning methods dominate the field. During this period, research on contracts remains limited, primarily due to the private and proprietary nature of these documents, which are not readily accessible online. As a result, relatively few studies are conducted compared to recent years. From 2020 to 2025, interest in the field increases significantly, driven by the release of publicly available contractual datasets, beginning with UNFAIR-ToS \citep{lippi2019claudette} and LEDGAR \citep{tuggener2020ledgar}. Since 2020, there has been a significant shift toward Transformer-based methods, which now dominate legal contract classification research. Notably, from 2023 onwards, these Transformer-based approaches become the standard, while traditional machine learning and deep learning techniques see a marked decline in usage.

\begin{figure*}[ht!]
\centering
\includegraphics[width=1\textwidth]{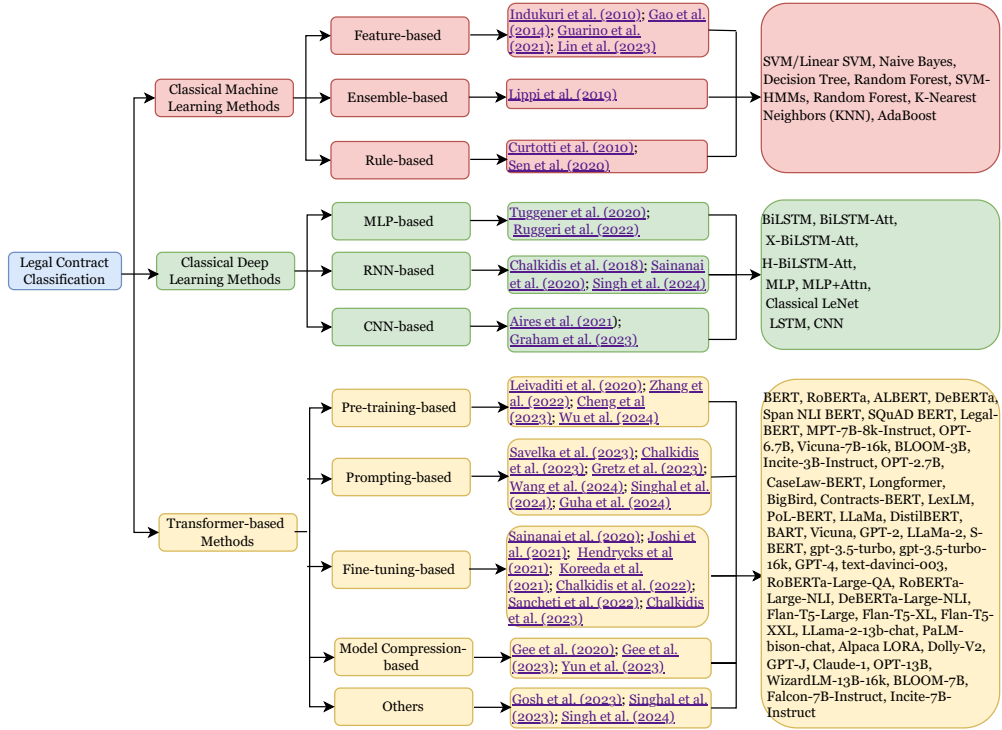}
\caption{Overview of Legal Contract Classification Methodology-based Taxonomy}\label{Fig3}
\end{figure*}

Figure \ref{Fig3} further illustrates a fine-grained, methodology-based taxonomy for legal contract classification, providing a comprehensive breakdown of the employed techniques. This taxonomy organizes the methods into a clear, structured framework that enhances the logical flow of the discussion. By doing so, it facilitates a deeper understanding of the evolution and current state of legal contract classification methodologies. Additionally, the taxonomy helps identify research gaps, guides future work, and enables systematic comparison of various techniques, highlighting their strengths, weaknesses, and applicability. As such, it serves both as an evaluation tool and a foundation for advancing research in legal contract classification.
It is important to note that this study focuses exclusively on methods related to legal contract classification tasks and datasets. While some of these methods may be applicable to other domains, such aspects fall outside the scope of this survey.
\begin{table*}[ht!]
\caption{Overview of Task-Specific Methodology}
\centering
\label{Task-Method-Mapping}
\resizebox{0.86\textwidth}{!}{
\begin{tabular}{|l|llllllll|lll|}
\hline
 &
  \multicolumn{8}{c|}{{\color[HTML]{3531FF} \textbf{Task}}} &
  \multicolumn{3}{c|}{{\color[HTML]{3531FF} \textbf{Method}}} \\ \cline{2-12} 
\multirow{-2}{*}{} &
  \multicolumn{1}{l|}{\cellcolor[HTML]{F8FF00}\rotatebox{90}{Topic Classification}} &
  \multicolumn{1}{l|}{\cellcolor[HTML]{FE996B}\rotatebox{90}{Risky/Unfair Clause Identification}} &
  \multicolumn{1}{l|}{\cellcolor[HTML]{FFC702}\rotatebox{90}{Deontic Modality Classification}} &
  \multicolumn{1}{l|}{\cellcolor[HTML]{38FFF8}\rotatebox{90}{Contractual Ambiguity Identification}} &
  \multicolumn{1}{l|}{\cellcolor[HTML]{FD6864}\rotatebox{90}{Norm Conflict Identification}} &
  \multicolumn{1}{l|}{\cellcolor[HTML]{CD9934}\rotatebox{90}{Obligatory Clause Classification}} &
  \multicolumn{1}{l|}{\cellcolor[HTML]{34FF34}\rotatebox{90}{NLI for Contracts}} &
  \cellcolor[HTML]{FE0000}\rotatebox{90}{Others} &
  \multicolumn{1}{l|}{\cellcolor[HTML]{68CBD0}\rotatebox{90}{Classical Machine Learning}} &
  \multicolumn{1}{l|}{\cellcolor[HTML]{C0C0C0}\rotatebox{90}{Classical Deep Learning}} &
  \cellcolor[HTML]{9698ED}\rotatebox{90}{Transformer-based} \\ \hline
\citep{indukuri2010mining} &
  \multicolumn{1}{l|}{} &
  \multicolumn{1}{l|}{} &
  \multicolumn{1}{l|}{} &
  \multicolumn{1}{l|}{} &
  \multicolumn{1}{l|}{} &
  \multicolumn{1}{l|}{\cellcolor[HTML]{CD9934}} &
  \multicolumn{1}{l|}{} &
   &
  \multicolumn{1}{l|}{\cellcolor[HTML]{68CBD0}{\color[HTML]{68CBD0} }} &
  \multicolumn{1}{l|}{} &
   \\ \hline
\citep{curtotti2010corpus} &
  \multicolumn{1}{l|}{} &
  \multicolumn{1}{l|}{} &
  \multicolumn{1}{l|}{} &
  \multicolumn{1}{l|}{} &
  \multicolumn{1}{l|}{} &
  \multicolumn{1}{l|}{} &
  \multicolumn{1}{l|}{} &
  \cellcolor[HTML]{FE0000}{\color[HTML]{009901} } &
  \multicolumn{1}{l|}{\cellcolor[HTML]{68CBD0}{\color[HTML]{68CBD0} }} &
  \multicolumn{1}{l|}{} &
   \\ \hline
\citep{gao2014extracting} &
  \multicolumn{1}{l|}{} &
  \multicolumn{1}{l|}{} &
  \multicolumn{1}{l|}{\cellcolor[HTML]{FFC702}} &
  \multicolumn{1}{l|}{} &
  \multicolumn{1}{l|}{} &
  \multicolumn{1}{l|}{} &
  \multicolumn{1}{l|}{} &
   &
  \multicolumn{1}{l|}{\cellcolor[HTML]{68CBD0}{\color[HTML]{68CBD0} }} &
  \multicolumn{1}{l|}{} &
   \\ \hline
\citep{chalkidis2018obligation} &
  \multicolumn{1}{l|}{} &
  \multicolumn{1}{l|}{} &
  \multicolumn{1}{l|}{\cellcolor[HTML]{FFC702}} &
  \multicolumn{1}{l|}{} &
  \multicolumn{1}{l|}{} &
  \multicolumn{1}{l|}{} &
  \multicolumn{1}{l|}{} &
   &
  \multicolumn{1}{l|}{} &
  \multicolumn{1}{l|}{\cellcolor[HTML]{C0C0C0}} &
   \\ \hline
\citep{lippi2019claudette} &
  \multicolumn{1}{l|}{} &
  \multicolumn{1}{l|}{\cellcolor[HTML]{FE996B}{\color[HTML]{FFCE93} }} &
  \multicolumn{1}{l|}{} &
  \multicolumn{1}{l|}{} &
  \multicolumn{1}{l|}{} &
  \multicolumn{1}{l|}{} &
  \multicolumn{1}{l|}{} &
   &
  \multicolumn{1}{l|}{\cellcolor[HTML]{68CBD0}{\color[HTML]{68CBD0} }} &
  \multicolumn{1}{l|}{\cellcolor[HTML]{C0C0C0}} &
   \\ \hline
\citep{tuggener2020ledgar} &
  \multicolumn{1}{l|}{\cellcolor[HTML]{F8FF00}} &
  \multicolumn{1}{l|}{} &
  \multicolumn{1}{l|}{} &
  \multicolumn{1}{l|}{} &
  \multicolumn{1}{l|}{} &
  \multicolumn{1}{l|}{} &
  \multicolumn{1}{l|}{} &
   &
  \multicolumn{1}{l|}{\cellcolor[HTML]{68CBD0}} &
  \multicolumn{1}{l|}{\cellcolor[HTML]{C0C0C0}} &
  \cellcolor[HTML]{9698ED} \\ \hline
\citep{leivaditi2020benchmark} &
  \multicolumn{1}{l|}{} &
  \multicolumn{1}{l|}{\cellcolor[HTML]{FE996B}{\color[HTML]{FFCE93} }} &
  \multicolumn{1}{l|}{} &
  \multicolumn{1}{l|}{} &
  \multicolumn{1}{l|}{} &
  \multicolumn{1}{l|}{} &
  \multicolumn{1}{l|}{} &
   &
  \multicolumn{1}{l|}{} &
  \multicolumn{1}{l|}{} &
  \cellcolor[HTML]{9698ED} \\ \hline
\citep{sainani2020extracting} &
  \multicolumn{1}{l|}{} &
  \multicolumn{1}{l|}{} &
  \multicolumn{1}{l|}{} &
  \multicolumn{1}{l|}{} &
  \multicolumn{1}{l|}{} &
  \multicolumn{1}{l|}{\cellcolor[HTML]{CD9934}} &
  \multicolumn{1}{l|}{} &
   &
  \multicolumn{1}{l|}{\cellcolor[HTML]{68CBD0}} &
  \multicolumn{1}{l|}{\cellcolor[HTML]{C0C0C0}} &
  \cellcolor[HTML]{9698ED} \\ \hline
\citep{sen2020learning} &
  \multicolumn{1}{l|}{} &
  \multicolumn{1}{l|}{} &
  \multicolumn{1}{l|}{} &
  \multicolumn{1}{l|}{} &
  \multicolumn{1}{l|}{} &
  \multicolumn{1}{l|}{\cellcolor[HTML]{CD9934}} &
  \multicolumn{1}{l|}{} &
   &
  \multicolumn{1}{l|}{\cellcolor[HTML]{68CBD0}} &
  \multicolumn{1}{l|}{\cellcolor[HTML]{C0C0C0}} &
   \\ \hline
\citep{guarino2021machine} &
  \multicolumn{1}{l|}{} &
  \multicolumn{1}{l|}{\cellcolor[HTML]{FE996B}} &
  \multicolumn{1}{l|}{} &
  \multicolumn{1}{l|}{} &
  \multicolumn{1}{l|}{} &
  \multicolumn{1}{l|}{} &
  \multicolumn{1}{l|}{} &
   &
  \multicolumn{1}{l|}{\cellcolor[HTML]{68CBD0}} &
  \multicolumn{1}{l|}{\cellcolor[HTML]{C0C0C0}} &
   \\ \hline
\citep{aires2021norm} &
  \multicolumn{1}{l|}{} &
  \multicolumn{1}{l|}{} &
  \multicolumn{1}{l|}{} &
  \multicolumn{1}{l|}{} &
  \multicolumn{1}{l|}{\cellcolor[HTML]{FD6864}} &
  \multicolumn{1}{l|}{} &
  \multicolumn{1}{l|}{} &
   &
  \multicolumn{1}{l|}{} &
  \multicolumn{1}{l|}{\cellcolor[HTML]{C0C0C0}} &
   \\ \hline
\citep{joshi2021domain} &
  \multicolumn{1}{l|}{} &
  \multicolumn{1}{l|}{} &
  \multicolumn{1}{l|}{\cellcolor[HTML]{FFC702}} &
  \multicolumn{1}{l|}{} &
  \multicolumn{1}{l|}{} &
  \multicolumn{1}{l|}{} &
  \multicolumn{1}{l|}{} &
   &
  \multicolumn{1}{l|}{\cellcolor[HTML]{68CBD0}} &
  \multicolumn{1}{l|}{\cellcolor[HTML]{C0C0C0}} &
  \cellcolor[HTML]{9698ED} \\ \hline
\citep{hendrycks2021cuad} &
  \multicolumn{1}{l|}{} &
  \multicolumn{1}{l|}{} &
  \multicolumn{1}{l|}{} &
  \multicolumn{1}{l|}{} &
  \multicolumn{1}{l|}{} &
  \multicolumn{1}{l|}{} &
  \multicolumn{1}{l|}{} &
  \cellcolor[HTML]{FE0000} &
  \multicolumn{1}{l|}{} &
  \multicolumn{1}{l|}{} &
  \cellcolor[HTML]{9698ED} \\ \hline
\citep{koreeda2021contractnli} &
  \multicolumn{1}{l|}{} &
  \multicolumn{1}{l|}{} &
  \multicolumn{1}{l|}{} &
  \multicolumn{1}{l|}{} &
  \multicolumn{1}{l|}{} &
  \multicolumn{1}{l|}{} &
  \multicolumn{1}{l|}{\cellcolor[HTML]{34FF34}} &
   &
  \multicolumn{1}{l|}{\cellcolor[HTML]{68CBD0}} &
  \multicolumn{1}{l|}{} &
  \cellcolor[HTML]{9698ED} \\ \hline
\citep{zhang2022task} &
  \multicolumn{1}{l|}{\cellcolor[HTML]{F8FF00}} &
  \multicolumn{1}{l|}{\cellcolor[HTML]{FE996B}} &
  \multicolumn{1}{l|}{} &
  \multicolumn{1}{l|}{} &
  \multicolumn{1}{l|}{} &
  \multicolumn{1}{l|}{} &
  \multicolumn{1}{l|}{} &
   &
  \multicolumn{1}{l|}{} &
  \multicolumn{1}{l|}{} &
  \cellcolor[HTML]{9698ED} \\ \hline
\citep{chalkidis2022lexglue} &
  \multicolumn{1}{l|}{\cellcolor[HTML]{F8FF00}} &
  \multicolumn{1}{l|}{\cellcolor[HTML]{FE996B}} &
  \multicolumn{1}{l|}{} &
  \multicolumn{1}{l|}{} &
  \multicolumn{1}{l|}{} &
  \multicolumn{1}{l|}{} &
  \multicolumn{1}{l|}{} &
   &
  \multicolumn{1}{l|}{\cellcolor[HTML]{68CBD0}} &
  \multicolumn{1}{l|}{} &
  \cellcolor[HTML]{9698ED} \\ \hline
\citep{sancheti2022agent} &
  \multicolumn{1}{l|}{} &
  \multicolumn{1}{l|}{} &
  \multicolumn{1}{l|}{\cellcolor[HTML]{FFC702}} &
  \multicolumn{1}{l|}{} &
  \multicolumn{1}{l|}{} &
  \multicolumn{1}{l|}{} &
  \multicolumn{1}{l|}{} &
   &
  \multicolumn{1}{l|}{\cellcolor[HTML]{68CBD0}} &
  \multicolumn{1}{l|}{} &
  \cellcolor[HTML]{9698ED} \\ \hline
\citep{ruggeri2022detecting} &
  \multicolumn{1}{l|}{} &
  \multicolumn{1}{l|}{\cellcolor[HTML]{FE996B}} &
  \multicolumn{1}{l|}{} &
  \multicolumn{1}{l|}{} &
  \multicolumn{1}{l|}{} &
  \multicolumn{1}{l|}{} &
  \multicolumn{1}{l|}{} &
   &
  \multicolumn{1}{l|}{\cellcolor[HTML]{68CBD0}} &
  \multicolumn{1}{l|}{\cellcolor[HTML]{C0C0C0}} &
   \\ \hline
\citep{gee2022fast} &
  \multicolumn{1}{l|}{\cellcolor[HTML]{F8FF00}} &
  \multicolumn{1}{l|}{} &
  \multicolumn{1}{l|}{} &
  \multicolumn{1}{l|}{} &
  \multicolumn{1}{l|}{} &
  \multicolumn{1}{l|}{} &
  \multicolumn{1}{l|}{} &
   &
  \multicolumn{1}{l|}{} &
  \multicolumn{1}{l|}{} &
  \cellcolor[HTML]{9698ED} \\ \hline
\citep{lin2023linear} &
  \multicolumn{1}{l|}{\cellcolor[HTML]{F8FF00}} &
  \multicolumn{1}{l|}{\cellcolor[HTML]{FE996B}} &
  \multicolumn{1}{l|}{} &
  \multicolumn{1}{l|}{} &
  \multicolumn{1}{l|}{} &
  \multicolumn{1}{l|}{} &
  \multicolumn{1}{l|}{} &
   &
  \multicolumn{1}{l|}{\cellcolor[HTML]{68CBD0}} &
  \multicolumn{1}{l|}{} &
   \\ \hline
\citep{graham2023natural} &
  \multicolumn{1}{l|}{} &
  \multicolumn{1}{l|}{} &
  \multicolumn{1}{l|}{\cellcolor[HTML]{FFC702}} &
  \multicolumn{1}{l|}{} &
  \multicolumn{1}{l|}{} &
  \multicolumn{1}{l|}{} &
  \multicolumn{1}{l|}{} &
   &
  \multicolumn{1}{l|}{\cellcolor[HTML]{68CBD0}} &
  \multicolumn{1}{l|}{\cellcolor[HTML]{C0C0C0}} &
   \\ \hline
\citep{cheng2023adapting} &
  \multicolumn{1}{l|}{} &
  \multicolumn{1}{l|}{\cellcolor[HTML]{FE996B}} &
  \multicolumn{1}{l|}{} &
  \multicolumn{1}{l|}{} &
  \multicolumn{1}{l|}{} &
  \multicolumn{1}{l|}{} &
  \multicolumn{1}{l|}{} &
   &
  \multicolumn{1}{l|}{} &
  \multicolumn{1}{l|}{} &
  \cellcolor[HTML]{9698ED} \\ \hline
\citep{chalkidis2023chatgpt} &
  \multicolumn{1}{l|}{\cellcolor[HTML]{F8FF00}} &
  \multicolumn{1}{l|}{\cellcolor[HTML]{FE996B}} &
  \multicolumn{1}{l|}{} &
  \multicolumn{1}{l|}{} &
  \multicolumn{1}{l|}{} &
  \multicolumn{1}{l|}{} &
  \multicolumn{1}{l|}{} &
   &
  \multicolumn{1}{l|}{} &
  \multicolumn{1}{l|}{} &
  \cellcolor[HTML]{9698ED} \\ \hline
\citep{gretz2023zero} &
  \multicolumn{1}{l|}{\cellcolor[HTML]{F8FF00}} &
  \multicolumn{1}{l|}{\cellcolor[HTML]{FE996B}} &
  \multicolumn{1}{l|}{} &
  \multicolumn{1}{l|}{} &
  \multicolumn{1}{l|}{} &
  \multicolumn{1}{l|}{} &
  \multicolumn{1}{l|}{\cellcolor[HTML]{34FF34}} &
  \cellcolor[HTML]{FE0000} &
  \multicolumn{1}{l|}{} &
  \multicolumn{1}{l|}{} &
  \cellcolor[HTML]{9698ED} \\ \hline
\citep{savelka2023unreasonable} &
  \multicolumn{1}{l|}{} &
  \multicolumn{1}{l|}{} &
  \multicolumn{1}{l|}{} &
  \multicolumn{1}{l|}{} &
  \multicolumn{1}{l|}{} &
  \multicolumn{1}{l|}{} &
  \multicolumn{1}{l|}{} &
  \cellcolor[HTML]{FE0000}{\color[HTML]{FE0000} } &
  \multicolumn{1}{l|}{} &
  \multicolumn{1}{l|}{} &
  \cellcolor[HTML]{9698ED} \\ \hline
\citep{chalkidis2023lexfiles} &
  \multicolumn{1}{l|}{\cellcolor[HTML]{F8FF00}{\color[HTML]{FFFC9E} }} &
  \multicolumn{1}{l|}{} &
  \multicolumn{1}{l|}{} &
  \multicolumn{1}{l|}{} &
  \multicolumn{1}{l|}{} &
  \multicolumn{1}{l|}{} &
  \multicolumn{1}{l|}{\cellcolor[HTML]{34FF34}} &
   &
  \multicolumn{1}{l|}{} &
  \multicolumn{1}{l|}{} &
  \cellcolor[HTML]{9698ED} \\ \hline
\citep{gee-etal-2023-multi} &
  \multicolumn{1}{l|}{\cellcolor[HTML]{F8FF00}{\color[HTML]{FFFC9E} }} &
  \multicolumn{1}{l|}{} &
  \multicolumn{1}{l|}{} &
  \multicolumn{1}{l|}{} &
  \multicolumn{1}{l|}{} &
  \multicolumn{1}{l|}{} &
  \multicolumn{1}{l|}{} &
   &
  \multicolumn{1}{l|}{} &
  \multicolumn{1}{l|}{} &
  \cellcolor[HTML]{9698ED} \\ \hline
\citep{yun2023focus} &
  \multicolumn{1}{l|}{\cellcolor[HTML]{F8FF00}{\color[HTML]{FFFC9E} }} &
  \multicolumn{1}{l|}{} &
  \multicolumn{1}{l|}{} &
  \multicolumn{1}{l|}{} &
  \multicolumn{1}{l|}{} &
  \multicolumn{1}{l|}{} &
  \multicolumn{1}{l|}{} &
   &
  \multicolumn{1}{l|}{} &
  \multicolumn{1}{l|}{} &
  \cellcolor[HTML]{9698ED} \\ \hline
\citep{ghosh-etal-2023-dale} &
  \multicolumn{1}{l|}{\cellcolor[HTML]{F8FF00}} &
  \multicolumn{1}{l|}{\cellcolor[HTML]{FE996B}} &
  \multicolumn{1}{l|}{} &
  \multicolumn{1}{l|}{} &
  \multicolumn{1}{l|}{} &
  \multicolumn{1}{l|}{} &
  \multicolumn{1}{l|}{} &
   &
  \multicolumn{1}{l|}{} &
  \multicolumn{1}{l|}{} &
  \cellcolor[HTML]{9698ED} \\ \hline
\citep{singhal2023towards} &
  \multicolumn{1}{l|}{} &
  \multicolumn{1}{l|}{\cellcolor[HTML]{FE996B}} &
  \multicolumn{1}{l|}{} &
  \multicolumn{1}{l|}{} &
  \multicolumn{1}{l|}{} &
  \multicolumn{1}{l|}{} &
  \multicolumn{1}{l|}{} &
   &
  \multicolumn{1}{l|}{} &
  \multicolumn{1}{l|}{} &
  \cellcolor[HTML]{9698ED} \\ \hline
\citep{singh2024data} &
  \multicolumn{1}{l|}{} &
  \multicolumn{1}{l|}{} &
  \multicolumn{1}{l|}{} &
  \multicolumn{1}{l|}{} &
  \multicolumn{1}{l|}{} &
  \multicolumn{1}{l|}{\cellcolor[HTML]{CD9934}} &
  \multicolumn{1}{l|}{} &
   &
  \multicolumn{1}{l|}{\cellcolor[HTML]{68CBD0}} &
  \multicolumn{1}{l|}{\cellcolor[HTML]{C0C0C0}} &
  \cellcolor[HTML]{9698ED} \\ \hline
\citep{wang2024metacognitive} &
  \multicolumn{1}{l|}{\cellcolor[HTML]{F8FF00}} &
  \multicolumn{1}{l|}{\cellcolor[HTML]{FE996B}} &
  \multicolumn{1}{l|}{} &
  \multicolumn{1}{l|}{} &
  \multicolumn{1}{l|}{} &
  \multicolumn{1}{l|}{} &
  \multicolumn{1}{l|}{} &
   &
  \multicolumn{1}{l|}{} &
  \multicolumn{1}{l|}{} &
  \cellcolor[HTML]{9698ED} \\ \hline
\citep{singhal2024generating} &
  \multicolumn{1}{l|}{} &
  \multicolumn{1}{l|}{} &
  \multicolumn{1}{l|}{} &
  \multicolumn{1}{l|}{\cellcolor[HTML]{38FFF8}} &
  \multicolumn{1}{l|}{} &
  \multicolumn{1}{l|}{} &
  \multicolumn{1}{l|}{} &
   &
  \multicolumn{1}{l|}{} &
  \multicolumn{1}{l|}{} &
  \cellcolor[HTML]{9698ED} \\ \hline
\citep{guha2024legalbench} &
  \multicolumn{1}{l|}{} &
  \multicolumn{1}{l|}{\cellcolor[HTML]{FE996B}} &
  \multicolumn{1}{l|}{} &
  \multicolumn{1}{l|}{} &
  \multicolumn{1}{l|}{} &
  \multicolumn{1}{l|}{} &
  \multicolumn{1}{l|}{\cellcolor[HTML]{34FF34}} &
  \cellcolor[HTML]{FE0000} &
  \multicolumn{1}{l|}{} &
  \multicolumn{1}{l|}{} &
  \cellcolor[HTML]{9698ED} \\ \hline
\citep{wu2024llama} &
  \multicolumn{1}{l|}{} &
  \multicolumn{1}{l|}{\cellcolor[HTML]{FE996B}} &
  \multicolumn{1}{l|}{} &
  \multicolumn{1}{l|}{} &
  \multicolumn{1}{l|}{} &
  \multicolumn{1}{l|}{} &
  \multicolumn{1}{l|}{} &
   &
  \multicolumn{1}{l|}{} &
  \multicolumn{1}{l|}{} &
  \cellcolor[HTML]{9698ED} \\ \hline
\end{tabular}
}
\end{table*}

\subsection{Classical Machine Learning Methods}
Classical methods characterize the feature-based approaches employed to automate the legal contract classification process, improving both accuracy and efficiency. The process generally begins with pre-processing, which involves tasks such as word segmentation (e.g., tokenization and stemming), data cleaning (removing stop words, special characters, and correcting spelling), and statistical analysis (e.g., frequency distribution and word co-occurrence). These steps lay the groundwork for applying various text representation techniques, such as Bag-of-Words (BOW), N-grams, TF-IDF, word2vec, and GloVe. BOW represents text as a vector of word frequencies, while N-grams capture adjacent word sequences to model contextual relationships. TF-IDF, on the other hand, assigns weights to words based on their frequency within a document and their rarity across the entire corpus, helping to highlight important terms. Word2vec and GloVe go a step further by generating dense word vectors that capture semantic relationships. While word2vec focuses on local context, GloVe uses both local and global statistics. Once the text has been represented appropriately, classifiers like Naive Bayes (NB), Support Vector Machines (SVM), and other machine learning models are employed for classification.

The summary of different methods and models used in classical machine learning methods is described in Table \ref{Table4}. These methods are widely applied in legal contract classification and continue to evolve. For instance, \cite{indukuri2010mining} utilize N-gram features with SVM to classify contract sentences into clauses and non-clauses, further distinguishing clauses based on their relevance to payment terms. Meanwhile, \cite{curtotti2010corpus} combine domain knowledge and linguistic rules in a hybrid feature approach, extracting 40 features using a hand-coded feature extractor. Their study experiments with several machine learning algorithms, including SVM, Naive Bayes, decision trees, and Random Forest (RF), along with rule-based and ensemble methods like Bagging and Majority Vote. The results show that integrating rule-based techniques with machine learning significantly improves classification performance. A notable approach introduced by \cite{sen2020learning} present RuleNN, a rule-based method that uses linguistic expressions (LEs) based on logical rules to classify sentences. These rules, expressed in first-order logic, are interpretable, allowing domain experts to verify and understand the model's decisions. For example, in the sentence "\emph{Notices may be transmitted electronically; by registered mail}", the LE identifies "\emph{transmit}" and "\emph{notice}", leading to the label "\emph{communication}". RuleNN outperforms other models, achieving Area Under the
Precision-Recall Curve (AUC-PR) scores 6.8×, 7.6×, and 1.5× higher than ILP, StarAI, and other neurosymbolic approaches, respectively. When compared to opaque models like BiLSTMs with GloVe embeddings, RuleNN’s use of LEs provides comparable performance but with the added advantage of explainability. \cite{gao2014extracting} explore linguistic features, such as phrasal features (e.g., modal phrases, main verbs) and contextual features (e.g., the use of "if" to indicate a clause), applying machine learning algorithms like Naive Bayes, SVM, and logistic regression. These techniques are used to extract six distinct classes of normative relationships from contracts. 

\cite{lippi2019claudette} employ an ensemble approach to detect potentially unfair clauses in contracts. Their method combines multiple models with different features: a single SVM with bag-of-words (unigrams, bigrams, and part-of-speech tags), eight SVMs for different unfairness categories, a single SVM using tree kernels (TK), an SVM-HMM for collective sentence classification, and eight SVM-HMMs for individual unfairness categories. They use a voting procedure, where sentences are classified as unfair if at least three models predict it. This ensemble approach outperforms single-feature-based machine learning algorithms and classical deep learning models like RNNs and CNNs, achieving the highest performance in detecting unfair clauses. \cite{guarino2021machine} introduce a sentence-based feature approach, utilizing the Google multilingual Universal Sentence Encoder (mUSE) to generate 512-dimensional sentence embeddings for each extracted clause. They use SVM, Random Forest (RF), K-Nearest Neighbors (KNN), and ensemble methods like AdaBoost (Ada) for classification. This approach surpasses state-of-the-art methods that rely on word-level features, such as bag-of-words. By leveraging sentence embeddings, the method captures broader context and meaning, leading to improved classification performance.

In recent studies, classical methods are often used as baselines for comparison with Transformer-based models like RoBERTa and BERT. For example, \cite{chalkidis2022lexglue} use a linear SVM as a baseline for comparing Transformer models. The SVM is trained on TF-IDF features extracted from the top-K most frequent n-grams (unigrams to trigrams) in the LexGLUE dataset. Other studies, such as \cite{sainani2020extracting} and \cite{leivaditi2020benchmark}, also use classical methods as baselines. However, with the superior performance of Transformer-based models, many researchers now prefer to train and deploy them directly for classification tasks.

Nonetheless, \cite{lin2023linear} argue that linear methods may still offer competitive results. Their research on the LexGLUE dataset shows that linear SVMs, employing the One-vs-Rest strategy, treating each label as a separate binary classification task, can be enhanced with techniques like thresholding and cost-sensitive learning. These methods remain appealing for their simplicity, scalability, and efficiency, especially when compared to the more complex Transformer models, which often require extensive hyper-parameter tuning.

\begin{table*}[ht!]
\centering
\caption{Summary of Classical Machine Learning Methods}
\label{Table4}
\resizebox{0.95\textwidth}{!}{
\begin{tabular}{lll}
\hline
\multicolumn{1}{c}{{\color[HTML]{FE0000} \textbf{Research Article}}} & \multicolumn{1}{c}{{\color[HTML]{FE0000} \textbf{Key Innovation}}} & \multicolumn{1}{c}{{\color[HTML]{FE0000} \textbf{Methods/Models}}} \\ \hline
\cite{indukuri2010mining} & N-grams features & SVM \\ \hline
\cite{curtotti2010corpus} & \begin{tabular}[c]{@{}l@{}}Hybrid feature approach \\ (40 features, hand-coded \\ feature extractor)\end{tabular} & \begin{tabular}[c]{@{}l@{}}ML: SVM, Naive Bayes, Decision Trees, \\ Cl. via Regression, Random Forest,  \\ Bagging, Majority Vote, Rule-based,\\ ML + Rule-based\end{tabular} \\ \hline
\cite{gao2014extracting} & \begin{tabular}[c]{@{}l@{}}Linguistic features (phrasal \\ and contextual features)\end{tabular} & \begin{tabular}[c]{@{}l@{}}Naive Bayes, SVM, logistic \\ regression, Hybrid of text patterns, \\ heuristics, and machine learning\end{tabular} \\ \hline
\cite{lippi2019claudette} & \begin{tabular}[c]{@{}l@{}}BoW (unigrams and bigrams \\ for words and part-of-speech \\ tags), tree kernels for \\ sentence representation, etc.\end{tabular} & \begin{tabular}[c]{@{}l@{}}Ensemble Methods, SVM, \\ SVM-HMMs, CNN, LSTM\end{tabular} \\ \hline
\cite{sen2020learning} & \begin{tabular}[c]{@{}l@{}}Rule-based method with \\linguistic features for \\ interpretability\end{tabular} & \begin{tabular}[c]{@{}l@{}}RuleNN, NeuralLP, BoostSRL (BSRL),\\ LSM, MITI, MIRI, metagol (MG), \\ MG$_{NT}$, MINet, BiLSTM\end{tabular} \\ \hline
\cite{guarino2021machine} & \begin{tabular}[c]{@{}l@{}}Sentence-based feature \\ approach using mUSE\end{tabular} & \begin{tabular}[c]{@{}l@{}}SVM, Random Forest, \\ K-Nearest Neighbors (KNN), and \\ ensemble methods like AdaBoost \\ (Ada)\end{tabular} \\ \hline
\cite{lin2023linear} & TF-IDF features & \begin{tabular}[c]{@{}l@{}}Linear SVM (One-vs-rest, \\ Thresholding, Cost-sensitive)\end{tabular} \\ \hline
\end{tabular}
}
\end{table*}

\subsection{Classical Deep Learning Methods}
In this section, classical deep learning methods are discussed, specifically those referring to pre-Transformer models for classification. These methods are divided into three categories: MLP-based, RNN-based, and CNN-based approaches. A summary of the different methods and models used in these classical approaches is provided in Table \ref{Table5}.

\subsubsection{MLP-based Approaches}
A Multi-Layer Perceptron (MLP) is a simple yet powerful neural network that models complex data through three layers: input, hidden, and output. The input layer represents data features, while the hidden layers use activation functions to capture non-linear patterns. The output layer generates predictions. During training, MLPs adjust their weights to improve accuracy. They excel in tasks like classification by learning complex data patterns. For instance, \cite{guarino2021machine} use an MLP-based model with mUSE to classify unfair clauses. \cite{tuggener2020ledgar} show that adding an attention layer to a BoW+MLP model outperforms traditional methods and Transformer-based models like DistilBERT in multi-label contract classification. Another study by \cite{ruggeri2022detecting} emphasizes the role of explainability in detecting unfair clauses in online Terms-of-Service agreements using a Memory-Augmented Neural Network (MANN). The MANN enhances traditional classification by leveraging external memory to store legal rationales. It computes the similarity between a clause and rationales using a two-layer MLP, retrieves relevant rationales from memory, and combines them with the clause for final classification. The model uses a sigmoid activation for attention, enabling the selection of multiple or no memory slots and updates the query through concatenation. This approach, tested on the Mmnet-ToS dataset, outperforms traditional methods like SVM, CNN, and LSTM.

\subsubsection{RNN-based Approaches}
Recurrent Neural Networks (RNNs) are widely used in text classification tasks due to their ability to capture long-term dependencies in sequential data. In these models, each word in the input is represented as a vector using word embeddings, and these vectors are processed sequentially through RNN cells, one at a time. The RNN captures the relationships between words across different time steps, maintaining shared parameters, which allows it to model context-dependent information effectively. The output from the final hidden layer is then used to predict the label for the input text, making RNNs ideal for tasks such as sentiment analysis and language modeling.

A significant advancement in this field came with the introduction of bidirectional RNNs, such as Bidirectional Long Short-Term Memory (BiLSTM) networks, which capture context from both past and future words in a sequence. This is particularly useful for tasks involving complex dependencies, such as legal clause classification. In this context, \cite{neill2017classifying} demonstrate that a BiLSTM classifier outperforms other classical methods, like logistic regression, SVM, AdaBoost, and Random Forests, especially in the task of legal clause classification. The BiLSTM's ability to model long-term dependencies, including modal verbs and negations, allows it to outperform methods that use fixed-size context windows.

Building upon these findings, \cite{chalkidis2018obligation} apply the BiLSTM classifier to legal contract classification, specifically using the Oblig \& Prohb dataset. They further enhance model performance by incorporating self-attention mechanisms, which enable the model to focus on important words within the input text. In addition, they explore a hierarchical BiLSTM architecture, which outperforms the flat BiLSTM by classifying clauses within the broader context of their discourse, rather than treating each clauses independently. This hierarchical approach, inspired by \cite{yang2016hierarchical}, is adapted to work at the sentence level, offering improvements over the document-level focus in the original work. Similarly, \cite{sainani2020extracting} and \cite{singh2024data} show that BiLSTM with attention outperforms classical methods such as SVM, Random Forest, and Naive Bayes in extracting obligation clauses from contracts. BiLSTM correctly identifies complex clauses as non-obligation by utilizing sequence-level information and attention, while the other methods misclassify them based on common obligation-related keywords.

\subsubsection{CNN-based Approaches} 
Convolutional Neural Networks (CNNs) leverage convolutional filters to extract features for image classification, and this approach is similarly applied in Natural Language Processing (NLP) tasks, such as text classification. In this context, input text is represented as a matrix of word vectors, which is then processed through convolutional layers using various filters, followed by pooling. The pooled features are combined into a final vector, which is subsequently used to predict the label for a given text. A study by \cite{aires2021norm} proposes a two-phase approach to detect potential conflicts between norms in contracts. The first phase involves identifying norms within contractual clauses using an SVM trained on a manually annotated dataset. In the second phase, a CNN is used to classify norm pairs as either conflicting or non-conflicting, further highlighting the effectiveness of CNNs in handling legal contractual text classification tasks. Similarly, \cite{graham2023natural} compares CNNs with classical methods like SVM, LR, and linear SVM with Stochastic Gradient Descent (SGD) for classifying norms and non-norms in contracts. The results demonstrate that CNNs outperform these traditional models, showcasing their superior ability to classify norms and non-norms effectively. Furthermore, in multilabel classification tasks, such as identifying deontic modalities in legal texts, CNNs also outperform classical and RNN-based approaches.

\begin{table*}[ht!]
\centering
\caption{Summary of Classical Deep Learning Methods}
\label{Table5}
\resizebox{0.95\textwidth}{!}{
\begin{tabular}{lll}
\hline
\multicolumn{1}{c}{{\color[HTML]{FE0000} \textbf{Research Article}}} & \multicolumn{1}{c}{{\color[HTML]{FE0000} \textbf{Key Innovation}}} & \multicolumn{1}{c}{{\color[HTML]{FE0000} \textbf{Models}}} \\ \hline
\cite{chalkidis2018obligation} & \begin{tabular}[c]{@{}l@{}}Concatenation of its word, \\ POS, shape embeddings\end{tabular} & \begin{tabular}[c]{@{}l@{}}BiLSTM, BiLSTM-Att,\\ X-BiLSTM-Att, \\ H-BiLSTM-Att\end{tabular} \\ \hline
\cite{sainani2020extracting} & TF-IDF (bigrams, trigrams) & \begin{tabular}[c]{@{}l@{}}BiLSTM-Att, SVM, \\ Random Forest, Naive\\ Bayes\end{tabular} \\ \hline
\cite{tuggener2020ledgar} & \begin{tabular}[c]{@{}l@{}}Unigram TFIDF, BoW,\\ Transformer-based \\ embeddings (DistilBERT)\end{tabular} & \begin{tabular}[c]{@{}l@{}}MLP, MLP+Attn, \\ Logistic Regression,\\ DistilBERT, Label name\end{tabular} \\ \hline
\cite{aires2021norm} & \begin{tabular}[c]
{@{}l@{}}Two-phase approach: SVM \\ for norm detection, CNN for\\ norm conflict classification.\end{tabular} & \begin{tabular}[c]{@{}l@{}}SVM, \\ Classical LeNet, CNN \end{tabular} \\ \hline
\cite{ruggeri2022detecting} & \begin{tabular}[c]{@{}l@{}}Memory-Augmented Neural \\ Network (MANN) explains \\ decision with rationales\end{tabular} & MANN, LSTM, CNN, SVM \\ \hline
\cite{graham2023natural} & \multicolumn{1}{c}{law2vec} & \begin{tabular}[c]{@{}l@{}}SVM, SVM+SGD training,\\ LR, NB, CNN, \\ CNN+law2vec, \\ LSTM+law2vec\end{tabular} \\ \hline
\cite{singh2024data} & TF-IDF (bigrams, trigrams) & \begin{tabular}[c]{@{}l@{}}BiLSTM-Att, SVM, \\ Random Forest, Naive\\ Bayes\end{tabular} \\ \hline
\end{tabular}
}
\end{table*}

\subsection{Transformer-based Methods}
This section discusses Transformer-based methods, categorized into five types: Pre-training-based, Prompting-based, Fine-tuning-based, Model Compression-based, and Miscellaneous approaches.
\subsubsection{Pre-training-based Approaches}
Pre-trained Transformer-based language models (PLMs) are trained on large, unsupervised corpora to learn fundamental language structures such as vocabulary, syntax, logic, and semantics. During pre-training, these models process extensive text data, including books, websites, and domain-specific documents, enabling them to develop a broad understanding of language. This general knowledge can later be fine-tuned for specific tasks, such as contract classification. PLMs typically use one of three architectures: encoder-based (e.g., BERT), decoder-based (e.g., GPT-2), or encoder-decoder (e.g., T5), with training objectives like autoregressive prediction, masked language modeling, or denoising tasks. A study by \cite{leivaditi2020benchmark} utilizes a general ALBERT model, pre-trained using Masked Language Modeling (MLM) on a corpus of lease agreements and fine-tuned for the red flag identification task. Their results show that ALBERT significantly improves when pre-trained on domain-specific data, highlighting the advantages of adapting the model to the specific language and features of lease contracts.

Recent research focuses on exploring the impact of these pre-training mechanisms. Below, we review some innovative approaches. One approach to improving the performance of PLMs in domain-specific tasks is to tailor the pre-training process to the target domain. Pre-training PLMs on raw domain-specific texts enhances domain knowledge but can sometimes significantly affect their prompting ability. To address this issue, \cite{cheng2023adapting} propose a method that adapts large language models (LLMs) by transforming raw domain-specific texts into reading comprehension tasks. Their approach automatically mines tasks like \emph{Summarization, Word-to-Text, Natural Language Inference (NLI), Commonsense Reasoning, Paragraph Detection, and Text Completion} from the domain corpus using regular expression regex-based patterns. These tasks are then used to pre-train the model in a self-supervised manner. By leveraging these diverse tasks, decoder-based models like GPT-J and LLaMA improve their understanding of domain-specific knowledge while maintaining strong performance in general prompting tasks. This method enhances performance on domain benchmarks such as LexGLUE for contractual language understanding.

Another promising approach to improving PLM performance is through multi-task pre-training. \cite{zhang2022task} introduce CompassMTL, a multi-task pre-training framework that combines both supervised and self-supervised objectives. Built on the DeBERTa architecture, CompassMTL incorporates supervised tasks, such as predicting the correct answer from multiple choices (e.g., question-answer matching), alongside self-supervised tasks like masked word prediction (similar to MLM). This dual approach leverages both labeled and unlabeled data, enhancing the model’s ability to generalize across different tasks. CompassMTL does not require changes to the underlying architecture and instead uses task-specific prefixes to differentiate between various tasks. Extensive experiments, including those on LexGLUE, show that CompassMTL contributes to improved performance, making it an effective and scalable approach.
A summary of the different methods and models used in these pre-training-based approaches is provided in Table \ref{Table6}.

\begin{table*}[ht!]
\centering
\caption{Summary of Pre-training-based Approaches}
\label{Table6}
\resizebox{0.8\textwidth}{!}{
\begin{tabular}{lll}
\hline
\multicolumn{1}{c}{{\color[HTML]{FE0000} \textbf{Research Article}}} & \multicolumn{1}{c}{{\color[HTML]{FE0000} \textbf{Key Innovation}}} & \multicolumn{1}{c}{{\color[HTML]{FE0000} \textbf{Models}}} \\ \hline
\cite{leivaditi2020benchmark} & \begin{tabular}[c]{@{}l@{}}Masked Language Modelling\\ pre-training\end{tabular} & ALBERT \\ \hline
\cite{zhang2022task} & \begin{tabular}[c]{@{}l@{}}Multi-task pre-training \\ framework\end{tabular} & DeBERTa \\ \hline
\cite{cheng2023adapting} & \begin{tabular}[c]{@{}l@{}}Adapts raw domain texts into \\ reading comprehension tasks\end{tabular} & \begin{tabular}[c]{@{}l@{}}GPT-J, \\ LLaMA\end{tabular} \\ \hline
\end{tabular}
}
\end{table*}
\subsubsection{Prompting-based Approaches}
\begin{figure*}[ht!]
\centering
\includegraphics[width=1\textwidth]{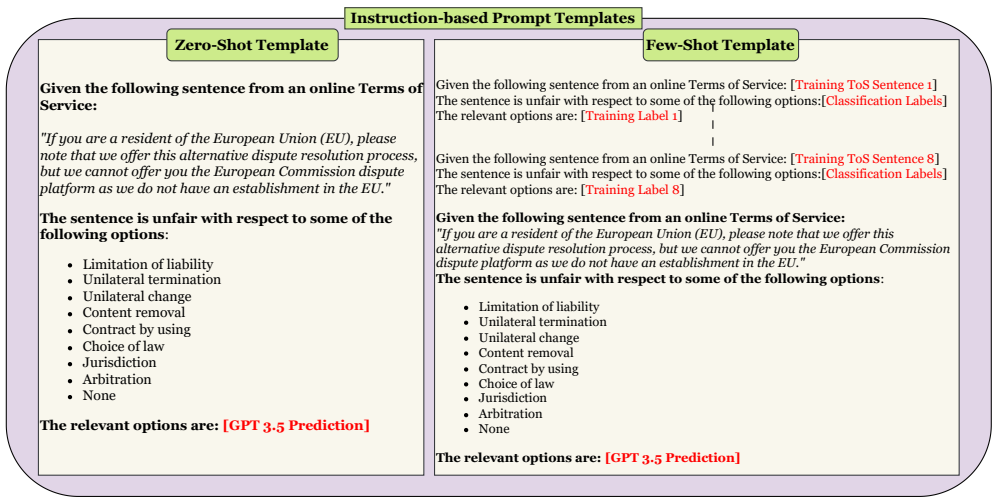}
\caption{An overview of the zero-shot and few-shot prompt templates designed by \citep{chalkidis2023chatgpt} for the task of unfair clause classification}\label{Promptingfig}
\end{figure*}

Prompting-based methods have become a popular approach in natural language processing (NLP), utilizing the capabilities of large pre-trained models through specially designed inputs. These methods involve crafting prompts to guide the model’s responses, either in zero-shot or few-shot learning scenarios. Recent studies highlight the growing significance of prompting-based methods in various contractual classification tasks. For example, \cite{chalkidis2023chatgpt} test template instruction-based prompting using GPT-3.5-turbo on the LexGLUE dataset (1k samples from UNFAIR-ToS and 10k from LEDGAR). In a zero-shot setting with the LEDGAR dataset, GPT-3.5-turbo demonstrates good performance, whereas its performance on UNFAIR-ToS is poor. However, in few-shot settings, where the model has access to eight training examples, performance deteriorates for LEDGAR but improves for UNFAIR-ToS. The zero-shot and few-shot instruction-based prompt templates used by \cite{chalkidis2023chatgpt} for the UNFAIR-ToS dataset, for example, are shown in Figure \ref{Promptingfig}. However, compared to fine-tuning-based approaches, the performance of zero-shot and few-shot instruction-based prompting using GPT-3.5-turbo remains lower on both tasks. This is because it is a general-purpose model without domain-specific fine-tuning.

\cite{gretz2023zero} evaluate the zero-shot performance of various models, including encoder-based models (S-BERT, RoBERTa-Large-QA, RoBERTa-Large-NLI, DeBERTa-Large-NLI) and encoder-decoder models (Flan-T5-Large, Flan-T5-XL, Flan-T5-XXL), using the TTC23 benchmark, which consists of 23 publicly available datasets from different domains, including contracts, such as LexGLUE, CUAD, and ContractNLI. The study finds that fine-tuning models like RoBERTa, DeBERTa, and Flan-T5-XXL on existing Topical Text Classification (TTC) datasets significantly improves their zero-shot performance when applied to new TTC datasets with different classes. 

\begin{figure*}[ht!]
\centering
\includegraphics[width=1\textwidth]{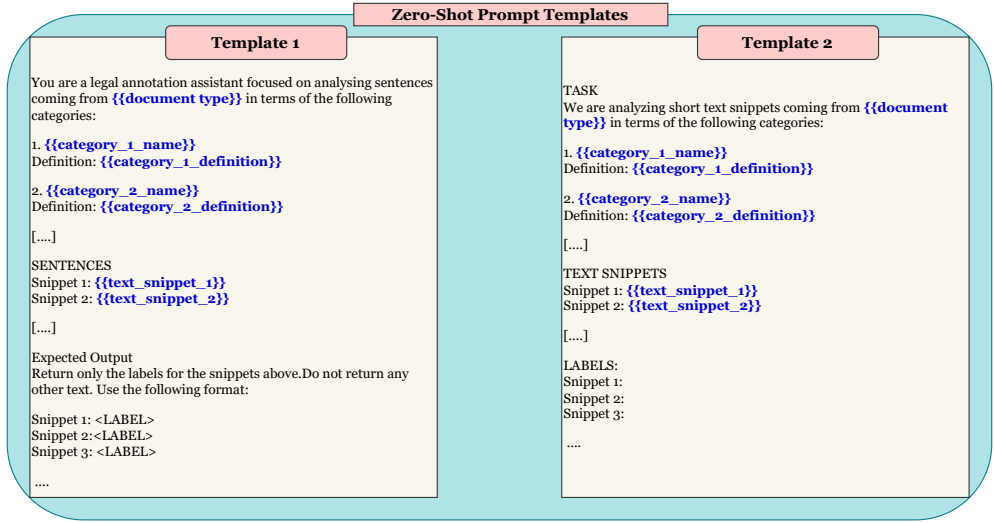}
\caption{Zero-shot prompt templates used by \cite{savelka2023unreasonable}. In both templates, placeholders were replaced as follows: ${{document\_type}}$ with “contract type of the CUAD”, ${{category\_n\_name}}$ with the specific label name, ${{category\_n\_definition}}$ with the label's definition, and ${{text\_snippet\_n}}$ with the contractual clause to be evaluated.}\label{Prompting1fig}
\end{figure*}

\cite{savelka2023unreasonable} select 3,783 clauses from the CUAD dataset, focusing on 12 common clause types, and compare the performance of zero-shot prompting using GPT models (gpt-3.5-turbo, text-davinci-003, gpt-3.5-turbo-16k, and GPT-4) with supervised models like RoBERTa and Random Forest. The best-performing model is RoBERTa, followed by GPT-4. Although GPT-4 is not trained on in-domain data, it performs similarly to a supervised Random Forest model. However, GPT-4 does not outperform a fine-tuned RoBERTa model, which is trained on thousands of task-specific examples, whereas GPT-4 has no access to such data. Two types of zero-shot prompt templates were used by \cite{savelka2023unreasonable}: \emph{Template 1} for models such as gpt-3.5-turbo, gpt-3.5-turbo-16k, and GPT-4, and \emph{Template 2} for the text-davinci-003 model, as illustrated in Figure \ref{Prompting1fig}.

\cite{wang2024metacognitive} introduce a novel prompting technique called \emph{Metacognitive Prompting (MP)}, which mimics human introspective reasoning. They test it on the LexGLUE dataset (600 samples from UNFAIR-ToS and 600 from LEDGAR). MP outperforms traditional prompting methods such as Zero-shot CoT \citep{kojima2022large} and Plan-and-Solve (PS) \citep{wang2023plan} in zero-shot settings, as well as Manual-CoT \citep{wei2022chain} and CoT-SC \citep{wangself} in few-shot settings. In evaluations using models such as LLaMA-2-13b-chat, PaLM-bison-chat, GPT-3.5-turbo, and GPT-4, MP with GPT-4 consistently demonstrates superior performance across most settings. Although the \cite{wang2024metacognitive} release prompts for several tasks, such as binary sentiment classification, similarity, paraphrasing, question answering, natural language inference, word sense disambiguation, and coreference resolution, they do not provide the prompts used for multi-label or multi-class classification tasks. \emph{Consequently, the specific prompts for topic classification using LEDGAR (multi-class) and unfair clause identification using UNFAIR-ToS (multi-label) remain unavailable.}

\begin{figure*}[ht!]
\centering
\includegraphics[width=0.9\textwidth]{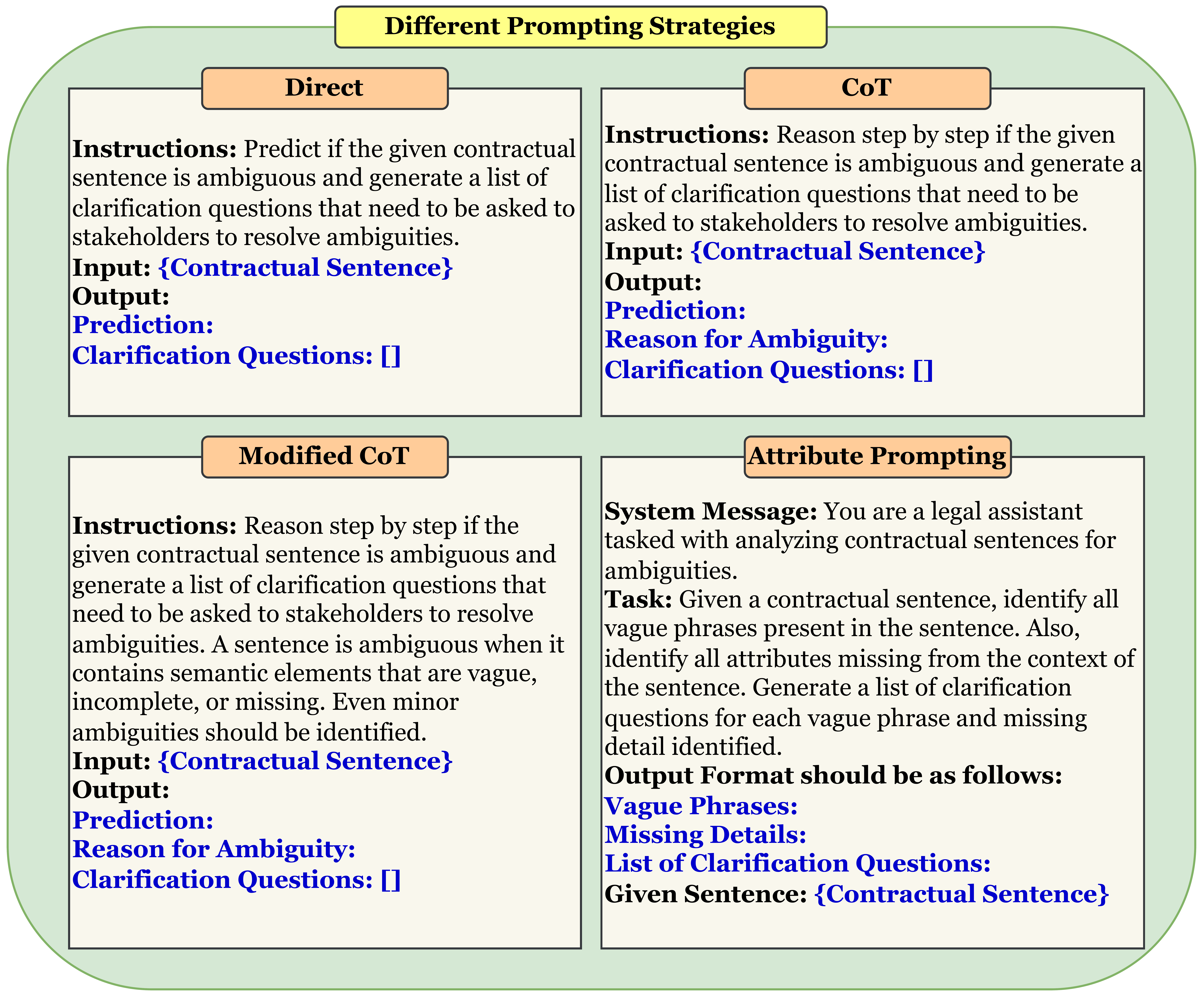}
\caption{Prompting strategies used for contractual ambiguity identification, as provided by \cite{singhal2024generating}, including Direct Prompting, Chain-of-Thought (CoT), Modified CoT, ConRAP, and ConRAP-Attribute Prompting.}\label{Prompting3fig}
\end{figure*}

\cite{singhal2024generating} introduce the ConRAP framework-a retrieval-based approach designed to tackle the challenge of identifying ambiguous terms in contractual clauses in a zero-shot setting. It employs a novel prompting technique called ConRAP-Attribute Prompting to detect vague or missing terms in contract clauses that create ambiguity, and generates clarification questions (CQs) to resolve these issues. After generating the CQs, ConRAP uses a retrieval-augmented question-answering (QA) method to search the entire contract for answers. If a CQ is already addressed in the contract, it is removed from the list. The remaining unanswered CQs highlight ambiguities that require further clarification. ConRAP outperforms other prompting techniques such as Direct Prompting, Chain-of-Thought (CoT), Modified CoT, and ConRAP-Attribute Prompting when used independently. The models used include ChatGPT (gpt-3.5-turbo), Vicuna, Alpaca-LoRA, and Dolly-V2, with ChatGPT (gpt-3.5-turbo) demonstrating the best overall performance. The different types of prompts used for this contractual ambiguity identification task are illustrated in Figure \ref{Prompting3fig}.

\cite{guha2024legalbench} evaluate 20 LLMs from 11 families, including GPT-3.5, GPT-4, and Claude-1, across various legal tasks. These tasks, designed to assess different aspects of legal reasoning, use the LEGALBENCH dataset for few-shot evaluation. Each task includes manually crafted prompts, some with 0 to 8 in-context examples to guide the models. The use of multiple models across these studies ensures a comprehensive evaluation of the proposed prompting techniques, demonstrating their effectiveness and versatility across various architectures and real-world scenarios. A summary of the different methods and models used in these prompting-based approaches is provided in Table \ref{Table7}.

\begin{table*}[ht!]
\centering
\caption{Summary of Prompting-based Approaches}
\label{Table7}
\resizebox{0.95\textwidth}{!}{
\begin{tabular}{lll}
\hline
\multicolumn{1}{c}{{\color[HTML]{FE0000} \textbf{Research Article}}} & \multicolumn{1}{c}{{\color[HTML]{FE0000} \textbf{Key Innovation}}} & \multicolumn{1}{c}{{\color[HTML]{FE0000} \textbf{Models}}} \\ \hline
\cite{chalkidis2023chatgpt} & \begin{tabular}[c]{@{}l@{}}Template instruction-based\\ Zero and Few-Shot Prompting\end{tabular} & GPT-3.5-Turbo \\ \hline
\cite{gretz2023zero} & \begin{tabular}[c]{@{}l@{}}Zero-Shot Prompting\\ \end{tabular} & \begin{tabular}[c]{@{}l@{}}S-BERT, RoBERTa-Large-QA, \\ RoBERTa-Large-NLI, \\ DeBERTa-Large-NLI, Flan-T5-Large,\\ Flan-T5-XL, Flan-T5-XXL\end{tabular} \\ \hline

\cite{savelka2023unreasonable} & \begin{tabular}[c]{@{}l@{}}Zero-Shot Prompting\\ \end{tabular} & \begin{tabular}[c]{@{}l@{}}gpt-3.5-turbo, gpt-3.5-turbo-16k, \\ GPT-4, text-davinci-003 model, \\ RoBERTa, Random Forest\end{tabular} \\ \hline

\cite{wang2024metacognitive} & \begin{tabular}[c]{@{}l@{}}Metacognitive Prompting (MP), \\ Zero-shot CoT, Plan-and-Solve\\ (PS), Manual-CoT, CoT-SC\end{tabular} & \begin{tabular}[c]{@{}l@{}}LLaMA-2-13b-chat, PaLM-bison-chat, \\ GPT-3.5-turbo, and GPT-4\end{tabular} \\ \hline
\cite{singhal2024generating} & \begin{tabular}[c]{@{}l@{}}ConRAP framework in a\\ Zero-Shot Setting\end{tabular} & \begin{tabular}[c]{@{}l@{}}ChatGPT, Vicuna, Alpaca LORA,\\ and Dolly-V2\end{tabular} \\ \hline
\cite{guha2024legalbench} & \begin{tabular}[c]{@{}l@{}}Zero and Few-Shot Prompting\\\end{tabular} & \begin{tabular}[c]{@{}l@{}}20 LLMs from 11 families, including\\ GPT-3.5, GPT-4, Claude-1, and others\end{tabular} \\ \hline
\end{tabular}
}
\end{table*}

\subsubsection{Fine-tuning-based Approaches}\label{sec5.3.3}
Supervised fine-tuning-based methods prove highly effective for legal contract classification (LCC) tasks in legal NLP. These methods leverage pre-trained Transformer models, which are then fine-tuned on task-specific (typically much smaller) labeled datasets. This specialized fine-tuning significantly enhances the model’s ability to perform tasks accurately, such as topic classification, identifying risky or unfair clauses, and more, while boosting both performance and efficiency. 

For example, \cite{sainani2020extracting} demonstrate that fine-tuning the encoder-based BERT model for classifying requirements from obligatory clauses in software engineering contracts results in better performance compared to classical and RNN-based methods. Similarly, \cite{joshi2021domain} show that domain adaptation, leveraging labeled regulations as training data due to their linguistic and taxonomical similarities with contracts, enables better classification of deontic modalities in contracts. They compare methods such as rule-based approaches, Bi-LSTMs, and BERT, with BERT outperforming the others in classifying deontic modalities. 

\cite{hendrycks2021cuad} further explore fine-tuning for key contract-clauses detection using extractive question answering. In this approach, models are trained with a question-answering framework, where the question consists of a label category and a brief description. The model then identifies relevant sections of the contract corresponding to each label. To manage long documents, PLMs like BERT, RoBERTa, ALBERT, and DeBERTa are fine-tuned with a sliding window technique, enhancing the model's accuracy in handling long-range contexts. This method significantly improves performance in legal contract classification tasks.  

Meanwhile, \cite{koreeda2021contractnli} present Span NLI BERT, a model combining evidence identification and natural language inference for contractual tasks. Unlike traditional models that predict start and end tokens, Span NLI BERT classifies spans using [SPAN] tokens within a multi-label binary classification framework, incorporating dynamic context splitting to ensure sufficient context for accurate span identification. Span NLI BERT outperforms several baselines, including Doc TF-IDF+SVM, Span TF-IDF+SVM, SQuAD BERT, and others. The authors find that increasing the model size enhances performance in both evidence identification and NLI. Furthermore, transferring DeBERTa-xlarge pretrained on CUAD \citep{hendrycks2021cuad} yields marginal gains in NLI, establishing it as the best-performing model on the ContractNLI dataset.

\cite{chalkidis2022lexglue} explore a hierarchical approach using multiple encoder-based models, including BERT, RoBERTa, DeBERTa, Legal-BERT, and CaseLaw-BERT, to process long legal texts from the LexGLUE dataset. This hierarchical method, similar to the one proposed by \cite{chalkidis-etal-2021-paragraph}, enhances the model’s handling of complex legal language. The study also investigates encoder-based models tailored for long texts, such as Longformer and BigBird, though these models do not employ the same hierarchical structure as the others mentioned. Legal-BERT demonstrates strong performance in most cases. 

\cite{sancheti2022agent} examine fine-tuning pre-trained language models (PLMs) for agent-specific multi-label deontic modality classification using the LEXDEMOD dataset. They compare three approaches: a majority class baseline, a rule-based method, and PLM fine-tuning (BERT, RoBERTa, ContractS-BERT). The study also examines how different training settings, like masking context or focusing on trigger spans, affect classification performance. 

\cite{chalkidis2023lexfiles} release two new legal PLMs, LexLM Base and Large, which are based on the RoBERTa architecture. These models are pre-trained on a multinational English legal corpus, LeXFiles. The authors fine-tune LexLM (Base and Large) on downstream tasks, including topic classification using the LEDGAR dataset from LexGLUE, and contract-based natural language inference using the ContractNLI dataset. They compare the performance of LexLM models with other PLMs such as RoBERTa, LegalBERT, CaseLaw-BERT, and PoL-BERT. Their results show that LexLM-Large outperforms other models on the topic classification task, while LegalBERT achieves the best performance on contract-based natural language inference.
 
\cite{wu2024llama} introduce "block expansion", a post-pretraining method that enhances off-the-shelf large language models (LLMs) by adding copied Transformer blocks initialized with zero weights in their linear layers, ensuring identity mapping at the start. These new blocks are tuned on a domain-specific corpus, while the original Transformer blocks remain frozen. This approach minimizes disruption to the pre-trained model while enabling targeted adaptation to specific tasks. After tuning, the extended model shows significant improvements in both general and domain-specific tasks. A summary of the different methods and models used in these fine-tuning-based approaches is provided in Table \ref{Table8}.

\begin{table*}[ht!]
\centering
\caption{Summary of Fine-tuning-based Approaches}
\label{Table8}
\resizebox{\textwidth}{!}{
\begin{tabular}{lll}
\hline
\multicolumn{1}{c}{{\color[HTML]{FE0000} \textbf{Research Article}}} & \multicolumn{1}{c}{{\color[HTML]{FE0000} \textbf{Key Innovation}}} & \multicolumn{1}{c}{{\color[HTML]{FE0000} \textbf{Models}}} \\ \hline
\cite{sainani2020extracting} & Simple Encoder-based Fine-tuning & \begin{tabular}[c]{@{}l@{}}BERT, BiLSTM, \\ Naive Bayes, SVM\\ Random Forest\end{tabular} \\ \hline
\cite{joshi2021domain} & Domain Adaptation Fine-tuning & \begin{tabular}[c]{@{}l@{}}BERT, BiLSTM, \\ Rule-based\end{tabular} \\ \hline
\cite{hendrycks2021cuad} & \begin{tabular}[c]{@{}l@{}}Fine-tuning with extractive question \\ answering and sliding window \\ technique\end{tabular} & \begin{tabular}[c]{@{}l@{}}BERT, RoBERTa, \\ ALBERT, DeBERTa\end{tabular} \\ \hline
\cite{koreeda2021contractnli} & \begin{tabular}[c]{@{}l@{}}Span NLI BERT combines evidence \\ identification and natural language \\ inference\end{tabular} & \begin{tabular}[c]{@{}l@{}}Span NLI BERT, \\ SQuAD BERT,\\ DeBERTa, SVM\end{tabular} \\ \hline
\cite{chalkidis2022lexglue} & \begin{tabular}[c]{@{}l@{}}Hierarchical approach for processing \\ long legal texts using multiple \\ encoder-based models\end{tabular} & \begin{tabular}[c]{@{}l@{}}BERT, RoBERTa, \\ DeBERTa, Legal-BERT, \\ CaseLaw-BERT, \\ Longformer, BigBird\end{tabular} \\ \hline
\cite{sancheti2022agent} & \begin{tabular}[c]{@{}l@{}}Fine-tuning for agent-specific multi-\\ label deontic modality classification\end{tabular} & \begin{tabular}[c]{@{}l@{}}BERT, RoBERTa, \\ Contracts-BERT, \\ Rule-based\end{tabular} \\ \hline
\cite{chalkidis2023lexfiles} & \begin{tabular}[c]{@{}l@{}}Fine-tuning LexLM models pre-\\ trained on the LeXFiles dataset for \\ downstream tasks\end{tabular} & \begin{tabular}[c]{@{}l@{}}LexLM (Base, Large), \\ RoBERTa, LegalBERT, \\ CaseLaw-BERT, \\ PoL-BERT\end{tabular} \\ \hline
\cite{wu2024llama} & \begin{tabular}[c]{@{}l@{}}Post-pretraining method "block \\ expansion" by adding copied \\ Transformer blocks\end{tabular} & LLaMA \\ \hline
\end{tabular}
}
\end{table*}
\subsubsection{Model Compression-based Approaches}
Recent advancements in PLMs, which now boast billions of parameters, have led to significant improvements in performance. However, their large size and high computational demands make them costly and difficult to deploy effectively. As a result, ongoing research focuses on optimizing these models to be smaller, faster, and more cost-efficient without sacrificing performance. Recent studies use domain-specific datasets, such as the LEDGAR dataset, to evaluate the performance of their methodologies, including distillation, token pruning, and other optimization techniques. While the primary focus is on demonstrating the generalizability of these frameworks, the positive results from legal datasets highlight their potential for future legal research applications and model optimization in legal contexts.

One such approach, proposed by \cite{gee-etal-2023-multi}, aims to reduce the computational cost of language models by using Multi-Word Tokenizers (MWTs). This method extends the tokenizer's vocabulary to include frequent multi-word expressions (n-grams), which are treated as single tokens. By reducing the length of text sequences and the overall token count, MWTs facilitate faster processing through early truncation, thereby improving model efficiency. In experiments with the LEDGAR dataset from LexGLUE, a 4-fold truncation of input sequences results in either comparable or improved performance, while achieving inference speedups of approximately 4.4x. If some performance degradation is acceptable, speedups can reach as high as 9.4x.

Similarly, \cite{gee2022fast} explore another technique called Vocabulary Transfer (VT), which adapts large language models to smaller, domain-specific tokenizers. VT works by transferring embedding knowledge from a general-purpose vocabulary to a specialized one, improving inference speed with minimal performance loss. Both methods proposed by \cite{gee-etal-2023-multi} and \cite{gee2022fast} highlight the significant role of tokenization in model compression. Both studies suggest that these approaches are compatible with traditional compression techniques like Knowledge Distillation (KD), and they conclude that combining these methods could further reduce model size and computational requirements while maintaining high performance. Similarly, \cite{yun2023focus} propose a method to optimize Transformer models by integrating token pruning and token combining. Token pruning uses fuzzy logic to eliminate less important tokens, mitigating mispruning risks, while token combining condenses input sequences to reduce model size. This approach enhances model performance, reduces memory costs, and is evaluated on the LEDGAR dataset from LexGLUE. A summary of the different methods and models used in these model compression-based approaches is provided in Table \ref{Table10}.

\begin{table*}[ht!]
\centering
\caption{Summary of Model Compression-based Approaches}
\label{Table10}
\resizebox{0.95\textwidth}{!}{
\begin{tabular}{lll}
\hline
\multicolumn{1}{c}{{\color[HTML]{FE0000} \textbf{Research Article}}} & \multicolumn{1}{c}{{\color[HTML]{FE0000} \textbf{Key Innovation}}} & \multicolumn{1}{c}{{\color[HTML]{FE0000} \textbf{Models}}} \\ \hline
\cite{gee2022fast} & \begin{tabular}[c]{@{}l@{}}Vocabulary Transfer (VT) method transfers embeddings \\ from general purpose to specialized vocabularies\end{tabular} & BERT$_{base}$ \\ \hline
\cite{gee-etal-2023-multi} & \begin{tabular}[c]{@{}l@{}}Multi-Word Tokenizers (MWTs) treating multi-word \\ expressions as single tokens\end{tabular} & \begin{tabular}[c]{@{}l@{}}BERT$_{base}$,\\ DistilBERT$_{base}$\end{tabular} \\ \hline
\cite{yun2023focus} & Token pruning and combining & BERT$_{base}$ \\ \hline
\end{tabular}
}
\end{table*}

\subsubsection{Miscellaneous Approaches}
The other novel methods not covered in the previous subsection are described here. A summary of these methods and models is provided in Table \ref{Table11}.

\emph{\textbf{Data Augmentation-based:}}
In classification tasks, particularly when dealing with imbalanced datasets, there is the challenge of insufficient labeled data for underrepresented classes, which can lead to biased or suboptimal performance. An effective way to address this challenge is through data augmentation, a technique that artificially expands the training set by adding additional natural or synthetic examples, helping to improve model performance and reduce bias.

One such technique, presented by \cite{ghosh-etal-2023-dale}, is DALE, a generative \textbf{D}ata \textbf{A}ugmentation framework designed for low-resource \textbf{LE}gal NLP. Legal documents, with their complex language and specialized vocabulary, require more than simple sentence rephrasing for effective data augmentation. DALE addresses this challenge by leveraging an Encoder-Decoder language model, BART, which is pre-trained on a large, unlabeled legal corpus using a novel denoising objective based on selective masking. Unlike traditional approaches that mask random entities, DALE selectively masks co-occurring and highly correlated spans of text, preserving critical legal structures. This encourages the model to learn general legal knowledge while avoiding overfitting to specific document details. This approach enables DALE to generate diverse, coherent, and semantically rich legal text augmentations, outperforming existing baselines in terms of coherence and complexity, as demonstrated on datasets like LexGLUE.

Another study by \cite{singhal2023towards} tackles the problem of identifying unfair clauses using self-training while also addressing the issues of class imbalance and limited labeled data through data augmentation. They use ChatGPT to generate additional clauses for the minority class ("\emph{clearly unfair}") based on a structured prompt. These clauses are reviewed by annotators and added to the training data. Then, they apply self-training, where a teacher model is trained on the labeled data, generates predictions for unlabeled clauses, and adds high-confidence predictions as pseudo-labels. This process continues with the teacher model being replaced by a student model until accuracy improvements stop. Both techniques improve the model's performance and generalization.

\emph{\textbf{Hybrid-based:}}  \cite{singh2024data} introduce a Data Decomposition-based Hierarchical (DDH) classification method aimed at automating the fine-grained, multi-label classification of contractual obligations. In the data decomposition phase, the dataset is first divided into smaller subsets, called "buckets". The formation of these buckets begins by embedding the obligation statements using DistilRoBERTa, followed by K-means clustering to group the statements into clusters. After clustering, the final label for each obligation statement, represented as a BF-R-CN triplet, is assigned to a bucket (ranging from B1 to Bk) based on the cluster containing the highest number of statements. For example, if the BF-R-CN triplet label is "\emph{Security-Compliance-Customers\_specific\_policy\_adherence}", the cluster with the most statements is selected. If Cluster 1 contains 10 sentences and Cluster 2 contains 20, the triplet label is assigned to the bucket of the cluster with the most statements (in this case, B2, since it contains 20 sentences).

Next, in the hierarchical classification phase, k+1 Transformer-based multi-label classifiers are employed, where k is the total number of clusters. In the first phase, a single Transformer-based multi-label classifier divides each statement in the testing dataset into bucket values ranging from B1 to Bk. In the second phase, k additional Transformer-based multi-label classifiers are used to further classify the obligation statements within each bucket into a triplet. The Transformer-based models used for the process are BERT, RoBERTa, and GPT-2. This hybrid approach, combining data decomposition and hierarchical classification, proves effective for the fine-grained classification of contractual obligations into their respective triplets.

\begin{table*}[ht!]
\centering
\caption{Summary of other novel Approaches}
\label{Table11}
\resizebox{0.95\textwidth}{!}{
\begin{tabular}{lll}
\hline
\multicolumn{1}{c}{{\color[HTML]{FE0000} \textbf{Research Article}}} & \multicolumn{1}{c}{{\color[HTML]{FE0000} \textbf{Key Innovation}}} & \multicolumn{1}{c}{{\color[HTML]{FE0000} \textbf{Models}}} \\ \hline
\cite{ghosh-etal-2023-dale} & \begin{tabular}[c]{@{}l@{}}DALE, a generative data augmentation framework \\ with selective masking to generate coherent and \\ diverse legal text augmentations\end{tabular} & BART$_{large}$ \\ \hline
\cite{singhal2023towards} & Self-training with data augmentation using ChatGPT & \begin{tabular}[c]{@{}l@{}}BERT, Vicuna,\\ LLaMA-2\end{tabular} \\ \hline
\cite{singh2024data} & \begin{tabular}[c]{@{}l@{}}Hybrid approach combining data decomposition \\ method and hierarchical classification\end{tabular} & \begin{tabular}[c]{@{}l@{}}BERT, RoBERTa,\\ GPT-2\end{tabular} \\ \hline
\end{tabular}
}
\end{table*}

\section{Evaluation Techniques and Results}\label{sec6}
In this section, we introduce a set of commonly used metrics in Section \ref{section7.1} to evaluate the performance of legal contract classification models, including accuracy, precision, recall, and F1-score. These metrics are among the most widely adopted for assessing legal contract classification tasks. As the complexity and specialization of legal contract classification increase, additional evaluation metrics are introduced in Sections \ref{section7.2} and \ref{section7.3} to provide a more detailed assessment of model performance. These include Macro-F1, Micro-F1, F2-score, balanced accuracy, mean Average Precision (mAP), Area Under the Precision-Recall Curve (AUC-PR), and Precision@X\% Recall. To illustrate progress in this area, we also present the best-achieved performance from previous legal contract classification works in Section \ref{sec7.4} and Table \ref{Table12}.

\subsection{Traditional Classification Metrics}\label{section7.1}
This section introduces the fundamental metrics used to evaluate legal contract classification models, which are essential for most tasks. Let $TP$, $FP$, $TN$, $FN$, and $N$ represent the counts of true positives, false positives, true negatives, false negatives, and total number of samples, respectively. These metrics include:

\emph{\textbf{Accuracy:}} Accuracy is one of the most fundamental and widely used evaluation metrics in legal contract classification (LCC) tasks, particularly when datasets are balanced or only slightly imbalanced. It represents the proportion of correctly classified samples and provides a straightforward measure of overall model performance. For instance, accuracy is used in clause classification by \cite{indukuri2010mining}, norm conflict identification by \cite{aires2021norm}, and deontic modality classification by \cite{sancheti2022agent}. Even when data imbalance exists, researchers often report accuracy alongside other metrics such as precision, recall, and F1-score to present a more comprehensive performance evaluation \citep{graham2023natural, aires2021norm, sancheti2022agent}. Due to its simplicity, accuracy remains a key metric in LCC research and continues to serve as a useful baseline for comparing model effectiveness. It is computed as:
\begin{center}
\emph{\textbf{Accuracy}} = $\frac{(TP + TN)}{N}$
\end{center}
\vspace{10pt}
\emph{\textbf{Precision, Recall, and F1-score:}} Precision, recall, and the F1-score are commonly preferred metrics over accuracy when dealing with imbalanced datasets, where one or more classes are underrepresented. Accuracy is misleading in such scenarios because a model that always predicts the majority class achieves high accuracy without effectively identifying the minority class. For example, \cite{guarino2021machine} use these metrics for unfair clause identification, where unfair clauses are underrepresented. Similarly, \cite{sancheti2022agent, joshi2021domain, graham2023natural} apply them in deontic modality classification tasks with class imbalance issues. Precision and recall emphasize the model’s ability to correctly identify minority class instances, while the F1-score provides a balanced summary of both. Therefore, these metrics provide a more meaningful and reliable evaluation of model performance in imbalanced classification tasks compared to accuracy.

\emph{\textbf{Precision:}} Precision is the proportion of correctly predicted positive instances out of all instances predicted as positive, computed as:
\begin{center}
\emph{\textbf{Precision}} = $\frac{TP}{(TP + FP)}$
\end{center}
\vspace{10pt}
\emph{\textbf{Recall:}} Recall is the proportion of correctly predicted positive instances out of all actual positive instances, computed as:
\begin{center}
\emph{\textbf{Recall}} = $\frac{TP}{(TP + FN)}$
\end{center}
\vspace{10pt}
\emph{\textbf{F1-score:}} The F1-score is the harmonic mean of precision and recall, providing a balanced measure of performance, ranging from 0 (worst) to 1 (best), computed as:
\begin{center}
\emph{\textbf{F1-score}} = $\frac{2 \times \text{Precision} \times \text{Recall}}{(\text{Precision} + \text{Recall})}$
\end{center}

\subsection{Advanced Classification Metrics}\label{section7.2}
Due to factors such as task formulation skews in the test set, advanced metrics are used. These include:

\emph{\textbf{Micro-F1 and Macro-F1:}} In multi-label classification scenarios, where each contractual sentence can have multiple labels, traditional metrics like accuracy, precision, recall, and F1 often prove insufficient. Unlike binary or multi-class classification, multi-label classification requires evaluating performance across multiple overlapping classes simultaneously. Consequently, metrics like Micro-F1 and Macro-F1 are preferred because they provide a more balanced and comprehensive evaluation. These metrics are widely used in legal contract classification (LCC) tasks, such as topic classification, unfair clause detection, and others, where multi-label assignments and class imbalance prevail. Therefore, Micro-F1 and Macro-F1 offer a more informative assessment of model performance than traditional metrics in such complex settings.

\emph{\textbf{Micro-F1:}} The Micro-F1 score calculates the overall precision and recall across all labels, treating each prediction equally, and is particularly useful for imbalanced data. It is computed as:
\begin{center}
\emph{\textbf{Micro-F1}} = $\frac{2 \times P_{\mu} \times R_{\mu}}{P_{\mu} + R_{\mu}}$
\end{center}
\vspace{10pt}
\begin{center}
$P_{\mu}$ = $\frac{\sum_{i=1}^{L} \text{TP}_i}{\sum_{i=1}^{L} (\text{TP}_i + \text{FP}_i)}$
$R_{\mu}$ = $\frac{\sum_{i=1}^{L} \text{TP}_i}{\sum_{i=1}^{L} (\text{TP}_i + \text{FN}_i)}$
\end{center}
\vspace{10pt}
where $L$ is the total number of labels, and $TP_i$, $FP_i$, and $FN_i$ are the True Positives, False Positives, and False Negatives for label $i$.

\emph{\textbf{Macro-F1:}} The Macro-F1 score treats each label equally, regardless of its frequency of occurrence, meaning it does not consider the distribution of labels in the dataset. This distinction makes the Macro-F1 score more sensitive to the performance on less frequent labels. It is computed as:
\begin{center}
\emph{\textbf{Macro-F1}} = $\frac{1}{L} \sum_{i=1}^{L} \text{F1}_i$
\end{center}
\vspace{10pt}
\begin{center}
$\text{F1}_i = \frac{2 \times P_i \times R_i}{P_i + R_i}$
$P_i = \frac{\text{TP}_i}{(\text{TP}_i + \text{FP}_i)}$
$R_i = \frac{\text{TP}_i}{(\text{TP}_i + \text{FN}_i)}$
\end{center}
\vspace{10pt}
where $L$ is the total number of labels, and $TP_i$, $FP_i$, and $FN_i$ are the True Positives, False Positives, and False Negatives for $i$-th abel.

\emph{\textbf{F2-score:}} The F2-score is a performance metric commonly used in classification tasks where recall is prioritized over precision. It is particularly useful when false negatives carry a higher cost than false positives, as in situations where minimizing false negatives is crucial. For example, \cite{singhal2024generating} used the F2-score for their task because false negatives, such as missing ambiguities, are more costly than false positives, which involve incorrectly labeling unambiguous cases as ambiguous. Since failure to detect ambiguities could have significant consequences, minimizing false negatives becomes the primary objective. The F2-score addresses this by placing more weight on recall. It is calculated as:
\begin{center}
\emph{\textbf{F2-score}} = $\frac{4 \times \text{Precision} \times \text{Recall}}{5 \times \text{Precision} + \text{Recall}}$
\end{center}
\vspace{10pt}

\subsection{Metrics for Specialized Tasks}\label{section7.3}
In addition to general classification metrics, specific metrics are employed to evaluate performance in specialized tasks such as question-answering (QA) and NLI. These include:

\emph{\textbf{Balanced Accuracy:}} Balanced accuracy is the arithmetic mean of sensitivity and specificity. It is particularly useful for imbalanced data, where one class is much more frequent than the other (e.g., 90\% negative and 10\% positive). In \cite{guha2024legalbench}, balanced accuracy ensures fair evaluation in a balanced binary clause classification task, where some categories are underrepresented. By accounting for class imbalance, balanced accuracy prevents bias toward more frequent categories and offers a more meaningful performance measure. It is computed as:
\begin{center}
\emph{\textbf{Balanced Accuracy}} = $\frac{1}{2} \left( \frac{TP}{TP + FN} + \frac{TN}{TN + FP} \right)$ = $\frac{(\text{Sensitivity} + \text{Specificity})}{2}$
\end{center}
\vspace{10pt}

\emph{\textbf{Precision@X\% Recall:}} Precision@X\% Recall is an evaluation metric that measures the model’s precision, how many of the predicted relevant clauses are truly relevant, at a fixed recall level. In the work by \cite{hendrycks2021cuad}, this metric is used to assess how accurately the model retrieves relevant clauses once it achieves a specified level of recall, such as 80\%. Recall indicates the proportion of all actual relevant clauses that the model correctly identifies, while precision reflects the relevance of the model’s predictions at that point. This is important in real-world scenarios like LCC, where one wants to balance the need for high recall (to capture most relevant clauses) while maintaining a reasonable level of precision (to minimize irrelevant clauses being flagged). Unlike aggregate metrics like F1 or AUC-PR, Precision@X\% Recall allows practitioners to evaluate performance at a specific operating point, offering clearer insight into the trade-off between precision and recall at that level. It is computed as:
\begin{center} 
\emph{\textbf{Precision at X\% Recall}} = $\frac{TP_t}{TP_t + FP_t}$ 
\end{center} 
\vspace{10pt}
where $TP_t$ is the number of true positives at the desired recall threshold, and $FP_t$ is the number of false positives at the desired recall threshold.

\emph{\textbf{Area Under the (Precision-Recall) Curve (AUC-PR):}} AUC-PR is particularly well-suited for evaluating model performance on imbalanced datasets, where traditional metrics like accuracy and ROC-AUC can be misleading. Unlike ROC-AUC, which includes true negatives (often abundant in imbalanced settings), AUC-PR focuses exclusively on the positive class by plotting precision against recall across varying thresholds. This makes it especially informative when the objective is to identify relevant but underrepresented instances. In the work by \cite{hendrycks2021cuad}, AUC-PR is used to evaluate how well a model distinguishes between relevant and irrelevant clauses across different thresholds. A higher AUC-PR value, closer to 1, indicates that the model performs well at identifying relevant clauses with high precision over a significant range of recall values. If the curve is steep, it suggests that the model maintains high precision even as recall increases, contributing to a higher AUC-PR. Conversely, an AUC-PR of 0.3 means that the model performs no better than random guessing, offering little to no value in distinguishing relevant from irrelevant clauses. It is computed as:

\begin{center}
\emph{\textbf{AUC-PR}} = $\int_0^1 P(R) \, dR$
\end{center}
\vspace{10pt}
where $P(R)$ is the precision as a function of recall $R$, and $dR$ is the differential change in recall.

\emph{\textbf{Mean Average Precision (MAP):}} Mean Average Precision (MAP) is a widely used evaluation metric for ranking tasks, especially effective in scenarios with imbalanced classes and a strong emphasis on high recall. In the context of risky clause identification \citep{leivaditi2020benchmark}, where only approximately 2.3\% of sentences are risky clauses, accuracy becomes misleading, a model that predicts only "non-risky" achieves over 97\%. The F1-score can also be misleading; a model may achieve a decent F1 while still missing many true red flags due to low recall. MAP is preferred because it evaluates how well the model ranks risky clauses higher than non-risky ones, making it ideal for imbalanced, high-recall tasks like this. A higher MAP indicates that the model consistently identifies most of the risky clauses. It is computed as:

\begin{center}
\emph{\textbf{Mean Average Precision (MAP)}} = $\frac{1}{Q} \sum_{q=1}^{Q} \text{AP}(q)$
\end{center}
\vspace{10pt}
\begin{center}
\emph{\textbf{Average Precision (AP)}} = $\frac{1}{R_q} \sum_{k=1}^{n_q} P_q(k) \cdot \text{Rel}_q(k)$
\end{center}
\vspace{10pt}
where $Q$ is the total number of queries, $\text{AP}(q)$ is the average precision for query $q$, $R_q$ is the number of relevant items in query $q$, $n_q$ is the number of ranked items, $P_q(k)$ is the precision at rank $k$, and $\text{Rel}_q(k)$ is an indicator function equal to 1 if the item at rank $k$ is relevant, 0 otherwise.

\subsection{Results}\label{sec7.4}
In Table \ref{Table12}, we present the best-achieved performance of each previous legal contract classification work. However, it is important to note that these results may not be directly comparable due to differences in datasets, evaluation metrics, and research objectives. For example, some studies, such as \cite{gee-etal-2023-multi}, focus on reducing the computational cost of language models while maintaining or even improving performance. Therefore, evaluating these studies solely based on performance improvement could lead to an unfair comparison. Similarly, \cite{wu2024llama} introduced a method primarily designed to assess generalizability across both general and domain-specific tasks, with legal contract classification being just one of several experiments. While the method performs well in contract classification, it is not the top-performing approach, as the primary focus of the study is not on maximizing performance in this area. As a result, direct comparisons are not feasible. Nevertheless, Table \ref{Table12} offers a general overview of the quantitative performance of legal contract classification methods, with all values reported as percentages. To complement this overview, we also provide a qualitative comparative analysis across modeling paradigms to identify broad trends and contextualize model effectiveness. Still, given the varying objectives, constraints, and data conditions of each study, direct comparison remains challenging.

For the topic classification task on the LEDGAR dataset from LexGLUE \citep{chalkidis2022lexglue}, transformer-based models consistently outperform classical methods. Among classical approaches, TF-IDF combined with SVM serves as a strong baseline \citep{lin2023linear, chalkidis2022lexglue}. However, BERT-based architectures, particularly models such as DeBERTa and CaseLaw-BERT, demonstrate superior performance, typically achieving gains of approximately 1-3\% in macro-F1 and micro-F1 scores compared to the TF-IDF combined with SVM baseline \citep{chalkidis2022lexglue, lin2023linear, zhang2022task}. Approaches that incorporate compression-based techniques into BERT-based architectures show competitive performance with trade-offs, often sacrificing 1-7\% in macro-F1 for greater computational efficiency \citep{gee-etal-2023-multi, gee2022fast, yun2023focus}. In contrast, prompting-based approaches, such as zero-shot prompting with models like GPT-3.5, achieve good performance in terms of micro-F1, achieving 70.1\%, but show substantially lower macro-F1 scores, achieving 56.7\%, primarily due to the lack of domain-specific fine-tuning \citep{chalkidis2023chatgpt}. In low-resource scenarios, data augmentation techniques such as DALE \citep{ghosh-etal-2023-dale} and metacognitive prompting \citep{wang2024metacognitive} show promising performance. However, \citep{wang2024metacognitive} also mentioned that LLMs like GPT-3.5 and GPT-4 often produce errors such as statutory misinterpretation and jurisprudential drift, indicating a tendency to misread legal texts and invent unsupported legal claims. These issues, reflecting challenges with legal language and reasoning, highlight the need for domain-specific adjustments when applying metacognitive prompting in legal tasks. \emph{In conclusion, for the topic classification task,  transformer-based models, especially BERT-based architectures such as DeBERTa, CaseLaw-BERT, DistilBERT, and RoBERTa, outperform classical baselines and offer a clear advantage in both performance and robustness across varying task conditions.}

For the unfair clause identification task on the UNFAIR-ToS dataset from LexGLUE \citep{chalkidis2022lexglue}, transformer-based models again consistently outperform classical approaches. Among traditional methods,  ensemble-based strategies, such as combinations of SVMs, tree kernels, and SVM-HMM \citep{lippi2019claudette}, and TF-IDF combined with SVM \citep{chalkidis2022lexglue} serve as strong baselines \citep{lippi2019claudette}. However, BERT-based architectures, particularly those pretrained on legal corpora such as Legal-BERT and CaseLaw-BERT, consistently achieve superior results \citep{chalkidis2022lexglue}. In addition, models such as CompassMTL, a multi-task learning framework based on DeBERTa and adapted to legal domains \citep{zhang2022task}, and methods like DAPT-6B, which adapt raw legal texts into a reading comprehension format with GPT-J-6B, also demonstrate competitive performance \citep{cheng2023adapting}. These domain-adapted transformer models typically yield improvements of 1–5\% in macro-F1 and micro-F1 scores over classical baselines. This performance gap underscores the value of pretraining on legal texts, enabling models to better capture the complex and nuanced language characteristic of legal documents, which enhances performance in unfair clause identification. In contrast, prompting-based approaches such as zero-shot and few-shot prompting with GPT-3.5 show significantly lower performance as compared to fine-tuning based models \citep{chalkidis2023chatgpt}. The results highlight the limitations of general-purpose models without task-specific fine-tuning. In low-resource scenarios, data augmentation methods such as DALE \citep{ghosh-etal-2023-dale} and metacognitive prompting \citep{wang2024metacognitive} show promising results. However, since they are evaluated on smaller datasets, direct comparisons with fully fine-tuned models remain difficult. \emph{In conclusion, for the unfair clause identification task benefits from pretraining on legal corpora. Transformer-based models that incorporate domain-specific knowledge, such as Legal-BERT, CaseLaw-BERT, DeBERTa-based Multi-task pre-training
framework, and DAPT-6B model demonstrate consistently superior performance. Similarly, for risky clause identification, ALBERT with additional pretraining outperforms both the non-pretrained model and classical approaches such as TF-IDF 2-grams combined with Random Forest.  These findings highlight the importance of domain adaptation in legal contract classification and underscore the effectiveness of leveraging legal-specific language patterns for complex highly imbalanced classification tasks.}

For the deontic modality and obligatory clause classification task, each study uses a different dataset, making it impossible to maintain a consistent dataset as done in the previous tasks. Additionally, evaluation metrics vary significantly across studies, so even general comparisons or measuring performance gains between methods are not feasible, as is evident from Table \ref{Table12}. Therefore, we present only a general model-based comparative performance analysis for this task. In this setting, \cite{gao2014extracting} finds that logistic regression performs best when compared to Naive Bayes and SVM. \cite{chalkidis2018obligation} shows that a Hierarchical BiLSTM with attention performs better than BiLSTM variants such as BiLSTM, BiLSTM-Att, and X-BiLSTM-Att. \cite{graham2023natural} reports that CNN achieves better performance than baselines including logistic regression, SVM, BiLSTM, Naive Bayes, SVM trained with stochastic gradient descent, and CNN combined with Law2Vec. Similarly, \cite{joshi2021domain} shows that BERT outperforms classical models such as BiLSTM and rule-based methods. Most recently, \cite{sancheti2022agent} finds that RoBERTa-large performs better than rule-based approaches, RoBERTa-base, BERT-base, and Contracts-BERT-base. \emph{In conclusion, for deontic modality classification if a transformer-based model is used for the study, particularly BERT-based architectures such as BERT and RoBERTa, it generally performs well compared to classical and rule-based approaches. This is the exact similar case for the obligatory clause classification. However, findings from \cite{sancheti2022agent} show that Contracts-BERT does not perform well for this task, indicating that pretraining on legal corpora does not always guarantee better results. The effectiveness of domain-specific models depends on the nature of the task, and in some cases, general-purpose transformer models also perform competitively.}

For the task of contractual ambiguity identification, there is currently only one study available in the literature. As a result, we evaluate the findings of that single study, \cite{singhal2024generating}. In this work, the authors propose a prompting-based retrieval framework called ConRAP-Retrieval, which operates in a zero-shot setting. It outperforms other prompting techniques such as Direct Prompting, Chain-of-Thought (CoT), Modified CoT, and their proposed ConRAP-Attribute Prompting when used independently without retrieval. Although the precision remains relatively low, the framework demonstrates strong performance in terms of recall and F2-score, which are crucial for identifying ambiguous clauses. While we acknowledge that prioritizing recall and F2-score is reasonable given the nature of the task, presenting accuracy and F1-score alongside these metrics in future work would strengthen the evaluation by offering a more balanced perspective. \emph{In conclusion, general prompting techniques such as Direct Prompting, CoT, and Modified CoT do not perform well for ambiguity resolution in legal contracts, as demonstrated by the findings of \cite{singhal2024generating}. This underscores the importance of task-adapted prompting strategies that incorporate domain knowledge, as general-purpose prompting appears insufficient for this specialized task.}

For the norm conflict identification task, two subtasks are addressed in the literature: norm identification and conflict detection \citep{aires2021norm,aires2017norm}. For norm identification, SVM performs well compared to other models such as Perceptron and Passive Aggressive. Although the Passive Aggressive model achieves the highest precision, SVM outperforms it in terms of recall and F1-score, which are more critical for this subtask, as they better reflect a model’s ability to comprehensively and accurately identify all relevant norms. For norm conflict identification, CNN is used and achieves competitive results. \emph{In conclusion, while current approaches demonstrate reasonable performance, particularly with classical models and CNN-based methods, exploring transformer-based architectures for norm conflict identification may offer significant improvements. Given their success in capturing complex contextual relationships in other legal contract classification tasks, as discussed above, transformer models could enhance both the accuracy and generalizability of conflict detection between norms.}

For the natural language inference (NLI) for contracts task, several studies explore the effectiveness of transformer-based models in identifying entailment relationships within contractual texts. \cite{koreeda2021contractnli} show that DeBERTa-xlarge, when pretrained on the CUAD dataset \citep{hendrycks2021cuad}, outperforms both classical models and other transformer-based models, including Doc TF-IDF + SVM, Span TF-IDF + SVM, SQuAD-BERT, and Span-NLI BERT. Similarly, \cite{chalkidis2023lexfiles} find that Legal-BERT achieves better performance compared to other legal-domain and general models such as RoBERTa, CaseLawBERT, PoL-BERT, and LexLM. Additionally, \cite{gretz2023zero} report that DeBERTa-large outperforms other strong baselines including S-BERT, RoBERTa, and Flan-T5. \emph{In conclusion, across multiple studies, transformer-based models consistently outperform classical approaches for the NLI task in the Contractual NLP domain. In particular, domain-adapted models such as DeBERTa-xlarge  pretrained on the CUAD dataset  and Legal-BERT demonstrate superior performance. These findings reinforce the effectiveness of large pretrained transformer architectures, especially when fine-tuned or adapted for legal data, in capturing complex entailment relations in contractual texts.}

\begin{table*}[ht!]
\centering
\caption{Summary of best-achieved performance of previous legal contract classification work, with all reported values presented as percentages}
\label{Table12}
\resizebox{\textwidth}{!}{
\begin{tabular}{|l|c|c|l|}
\hline
\multicolumn{1}{|c|}{{\color[HTML]{FE0000} \textbf{Research Article}}} &
  {\color[HTML]{FE0000} \textbf{Task}} &
  {\color[HTML]{FE0000} \textbf{\begin{tabular}[c]{@{}c@{}}Reported \\ Performance (\%)\end{tabular}}} &
  \multicolumn{1}{c|}{{\color[HTML]{FE0000} \textbf{Remark}}} \\ \hline
\citep{tuggener2020ledgar}     &  & \begin{tabular}[c]{@{}c@{}}Micro-P: 73, Macro-P: 72\\ Micro-R: 61, Macro-R: 69\\ Micro-F1: 67, Macro-F1: 71\end{tabular} &  \\ \cline{1-1} \cline{3-3}
\citep{zhang2022task}          &  & Micro-F1: 88.3, Macro-F1: 83.2                                                                                           &  \\ \cline{1-1} \cline{3-3}
\citep{chalkidis2022lexglue}   &  & Micro-F1: 88.3, Macro-F1: 83                                                                                             &  \\ \cline{1-1} \cline{3-3}
\citep{gee2022fast}            &  & F1: 81.03                                                                                                                &  \\ \cline{1-1} \cline{3-3}
\citep{lin2023linear}          &  & Micro-F1: 87.0, Macro-F1: 80.7                                                                                           &  \\ \cline{1-1} \cline{3-3}
\citep{chalkidis2023chatgpt}   &  & Micro-F1: 70.1, Macro-F1: 56.7                                                                                           &  \\ \cline{1-1} \cline{3-3}
\citep{gretz2023zero}          &  & Macro-F1: 55.86                                                                                                          &  \\ \cline{1-1} \cline{3-3}
\citep{chalkidis2023lexfiles}  &  & Micro-F1: 84.7, Macro-F1: 72.8                                                                                           &  \\ \cline{1-1} \cline{3-3}
\citep{gee-etal-2023-multi}    &  & Macro-F1: 82.12                                                                                                          &  \\ \cline{1-1} \cline{3-3}
\citep{yun2023focus}           &  & \begin{tabular}[c]{@{}c@{}}Micro-F1: 87.3, Macro-F1: 79.2\\ Acc: 87.3\end{tabular}                                       &  \\ \cline{1-1} \cline{3-3}
\citep{ghosh-etal-2023-dale}   &  & Micro-F1: 78.36                                                                                                          &  \\ \cline{1-1} \cline{3-3}
\citep{wang2024metacognitive} &
  \multirow{-18}{*}{\begin{tabular}[c]{@{}c@{}}Topic \\ Classification\end{tabular}} &
  Micro-F1: 78.1, Macro-F1: 62.8 &
  \multirow{-18}{*}{\begin{tabular}[c]{@{}l@{}}The topic classification task utilizes Micro-F1 \\ as the common metric, with performance \\ values ranging from 67\% to 88.3\%.\end{tabular}} \\ \hline
\citep{lippi2019claudette}     &  & \begin{tabular}[c]{@{}c@{}}Macro-P: 82.6, Macro-R: 79.7\\ Macro-F1: 80.5\end{tabular}                                    &  \\ \cline{1-1} \cline{3-3}
\citep{leivaditi2020benchmark} &  & MAP: 57.33, IP@R: 35.79                                                                                                  &  \\ \cline{1-1} \cline{3-3}
\citep{guarino2021machine}     &  & P: 90, R: 92, F1: 91                                                                                                     &  \\ \cline{1-1} \cline{3-3}
\citep{zhang2022task}          &  & Micro-F1: 96.3, Macro-F1: 84.3                                                                                           &  \\ \cline{1-1} \cline{3-3}
\citep{chalkidis2022lexglue}   &  & Micro-F1: 96.0, Macro-F1: 83                                                                                             &  \\ \cline{1-1} \cline{3-3}
\citep{ruggeri2022detecting}   &  & Macro-F1: 62.44                                                                                                          &  \\ \cline{1-1} \cline{3-3}
\citep{lin2023linear}          &  & Micro-F1: 95.4, Macro-F1: 80.3                                                                                           &  \\ \cline{1-1} \cline{3-3}
\citep{cheng2023adapting}      &  & Acc: 84.9                                                                                                                &  \\ \cline{1-1} \cline{3-3}
\citep{chalkidis2023chatgpt}   &  & Micro-F1: 64.7, Macro-F1: 32.5                                                                                           &  \\ \cline{1-1} \cline{3-3}
\citep{gretz2023zero}          &  & Macro-F1: 98.36                                                                                                          &  \\ \cline{1-1} \cline{3-3}
\citep{ghosh-etal-2023-dale}   &  & Micro-F1: 82.98                                                                                                          &  \\ \cline{1-1} \cline{3-3}
\citep{singhal2023towards}     &  & Acc: 84, Macro-F1: 74                                                                                                    &  \\ \cline{1-1} \cline{3-3}
\citep{wang2024metacognitive}  &  & Micro-F1: 75.6, Macro-F1: 55.8                                                                                           &  \\ \cline{1-1} \cline{3-3}
\citep{wu2024llama} &
  \multirow{-15}{*}{\begin{tabular}[c]{@{}c@{}}Risky/Unfair\\ Clause\\ Identification\end{tabular}} &
  Acc: 75.17 &
  \multirow{-14}{*}{\begin{tabular}[c]{@{}l@{}}The risky/unfair clause identification task \\ utilizes Micro-F1 as the common metric, \\ with performance values ranging from \\ 64.7\% to 96.3\%.\end{tabular}} \\ \hline
\citep{gao2014extracting}      &  & \begin{tabular}[c]{@{}c@{}}Weighted-P: 86, Weighted-R: 83\\ Weighted-F1: 84\end{tabular}                                 &  \\ \cline{1-1} \cline{3-3}
\citep{chalkidis2018obligation} &
   &
  \begin{tabular}[c]{@{}c@{}}Micro-P: 87, Macro-P: 95\\ Micro-R: 90, Macro-R: 95\\ Micro-F1: 89, Macro-F1: 95\\ Micro-AUC: 94, Macro-AUC: 98\end{tabular} &
   \\ \cline{1-1} \cline{3-3}
\citep{joshi2021domain}        &  & P: 90, R: 89.66                                                                                                          &  \\ \cline{1-1} \cline{3-3}
\citep{sancheti2022agent}      &  & \begin{tabular}[c]{@{}c@{}}P: 89.48, R: 89.21, F1: 92.42\\ Acc: 90.23\end{tabular}                                       &  \\ \cline{1-1} \cline{3-3}
\citep{graham2023natural} &
  \multirow{-12}{*}{\begin{tabular}[c]{@{}c@{}}Deontic \\ Modality \\ Classification\end{tabular}} &
  \multicolumn{1}{l|}{\begin{tabular}[c]{@{}l@{}}P: 89, R: 8, F1: 89, Acc: 88\\ P: 98, Acc: 90, Ranking loss: 2\end{tabular}} &
  \multirow{-11}{*}{\begin{tabular}[c]{@{}l@{}}The datasets and metrics used across each \\ research article for deontic modality \\ classification vary, making even general \\ comparisons difficult.\end{tabular}} \\ \hline
\citep{singhal2024generating} &
  \begin{tabular}[c]{@{}c@{}}Contractual\\ Ambiguity\\ Identification\end{tabular} &
  P: 64, R: 97, F2: 87 &
  \begin{tabular}[c]{@{}l@{}}The contractual ambiguity identification \\ task uses the F2 score as the metric, \\ resulting in a score of 87\%.\end{tabular} \\ \hline
\citep{aires2021norm} &
  \begin{tabular}[c]{@{}c@{}}Norm \\ Conflict\\ Identification\end{tabular} &
  \begin{tabular}[c]{@{}c@{}}P: 88, R: 94, F1: 91, Acc: 90\\ Acc: 84\end{tabular} &
  \begin{tabular}[c]{@{}l@{}}The norm conflict identification task uses\\  accuracy as the metric, with norm \\ identification accuracy at 90\% and norm \\ conflict identification accuracy at 84\%.\end{tabular} \\ \hline
\citep{indukuri2010mining}     &  & Acc: 79.58                                                                                                               &  \\ \cline{1-1} \cline{3-3}
\citep{sainani2020extracting}  &  & F1: 85.8                                                                                                                 &  \\ \cline{1-1} \cline{3-3}
\citep{sen2020learning}        &  & AUC-PR: 78.2                                                                                                             &  \\ \cline{1-1} \cline{3-3}
\citep{singh2024data} &
  \multirow{-4}{*}{\begin{tabular}[c]{@{}c@{}}Obligatory\\ Clause\\ Classification\end{tabular}} &
  \begin{tabular}[c]{@{}c@{}}Micro-P: 82, Micro-R: 60\\ Micro-F: 69\end{tabular} &
  \multirow{-4}{*}{\begin{tabular}[c]{@{}l@{}}The datasets and metrics used across each \\ research article for obligatory clause \\ classification vary, making even general \\ comparisons difficult.\end{tabular}} \\ \hline
\citep{koreeda2021contractnli} &  & Acc: 89.2,  F1(C): 40.5, F1(E): 85.9                                                                                     &  \\ \cline{1-1} \cline{3-3}
\citep{gretz2023zero}          &  & Macro-F1: 87.10                                                                                                          &  \\ \cline{1-1} \cline{3-3}
\citep{chalkidis2023lexfiles} &
  \multirow{-3}{*}{\begin{tabular}[c]{@{}c@{}}NLI for\\ Contracts\end{tabular}} &
  Micro-F1: 70.2, Macro-F1: 65.6 &
  \multirow{-3}{*}{\begin{tabular}[c]{@{}l@{}}The NLI for contracts uses Macro-F1 as\\ the common metric, with values ranging \\ from 65.6\% to 87.10\%.\end{tabular}} \\ \hline
\citep{curtotti2010corpus}     &  & F1: 87.76                                                                                                                &  \\ \cline{1-1} \cline{3-3}
\citep{hendrycks2021cuad}      &  & \begin{tabular}[c]{@{}c@{}}AUC-PR: 47.8, \\ Precision at 80\% Recall: 44.0\\ Precision at 90\% Recall: 17.8\end{tabular} &  \\ \cline{1-1} \cline{3-3}
\citep{gretz2023zero}          &  & Macro-F1: 96.71                                                                                                          &  \\ \cline{1-1} \cline{3-3}
\citep{savelka2023unreasonable} &
  \multirow{-6}{*}{Others} &
  \begin{tabular}[c]{@{}c@{}}Micro-P: 95, Micro-R: 95\\ Micro-F1: 95\end{tabular} &
  \multirow{-5}{*}{\begin{tabular}[c]{@{}l@{}}The datasets and metrics used across each \\ research article for different studies vary, \\ making general comparisons difficult.\end{tabular}} \\ \hline
\end{tabular}
}
\end{table*}

\section{Challenges and Future Directions}\label{sec8}
In this section, we discuss the primary challenges in legal contract classification and explore potential avenues for future advancements in this area.

\subsection{Challenges and Future Directions in terms of Datasets}

\emph{\textbf{Lack of Standard Benchmark Dataset for Contractual Language Understanding}}:  
A major challenge in the field of legal contract classification is the absence of a dedicated benchmark dataset specifically designed for understanding contractual language. While existing benchmarks, such as LexGLUE \citep{chalkidis2022lexglue}, contribute to the broader Legal NLP domain, they do not fully address the distinct complexities of contractual language. For instance, LexGLUE includes seven datasets, but only two of them are directly relevant to the Contractual NLP domain. Another resource, LEGALBENCH \citep{guha2024legalbench}, focuses on contract-related documents but is primarily developed for evaluating large language models in zero- and few-shot settings. Out of its 162 tasks, 125 tasks involve between 50 and 500 samples. This means it primarily supports lightweight tasks rather than comprehensive evaluation, as noted by \cite{niklaus2023lextreme}. This limitation hampers the ability to assess the true understanding of contractual language. Therefore, there is a critical need for a benchmark dataset that encompasses a variety of contractual documents and tasks, enabling a more thorough and accurate evaluation of models' capabilities in understanding the nuances of contractual language.

\emph{\textbf{Geographic and Jurisdictional Imbalance in Labeled Datasets}}: Most available labeled contract datasets focus primarily on the U.S. or EU, resulting in a significant lack of data from other countries and regions \citep{guha2024legalbench}. This geographic imbalance limits the ability of models to generalize across different legal systems and contract structures. Legal contracts vary not only in language and format but also in underlying legal principles, such as common law (e.g., UK, U.S., India), civil law (e.g., France, Germany), and hybrid systems (e.g., South Africa, China). For example, contract drafting in common law systems tends to be more detailed and precedent-based, while civil law systems emphasize statutory interpretation and often rely on standardized templates \citep{haapio2017contracts}. Additionally, countries adopt distinct commercial and regulatory frameworks. These differences affect not only the structure of legal contractual documents but also the expression of obligations and enforcement clauses within contracts \citep{osifo2025evolving}. As a result, there is a notable lack of benchmark datasets spanning multiple jurisdictions.

A promising direction is to include countries with shared legal foundations, such as the UK, India and others, for task like deontic modality classification, where similar legal reasoning patterns may apply. While proprietary datasets such as Contract Requirement \citep{sainani2020extracting} and Fine-Grained Obligation \citep{singh2024data} may help bridge some gaps, they are typically not publicly accessible. To improve the global applicability of legal contract classification models, it is essential to develop more diverse datasets that span multiple legal systems. Future research should focus on building cross-jurisdictional corpora that reflect variations in legal doctrines, document structures, and commercial norms.

\emph{\textbf{Lack of Transparent Annotation:}}
Annotated datasets for legal contract classification are often limited and lack transparency. Many studies mention expert annotators but fail to disclose their qualifications or the annotation process \citep{braun2024beg}. For unbiased and reliable NLP systems, it is essential to document both the methods and the annotator backgrounds. Transparency in these areas ensures fairness, builds trust, and supports effective system evaluation.

\emph{\textbf{Dataset Design, Quality, and Bias:}} The datasets discussed in Section \ref{sec5.2} provide valuable resources for legal contract classification but exhibit some limitations. LEDGAR \citep{tuggener2020ledgar} uses heuristic-based, semi-automatic labeling, which introduces annotation noise due to inconsistencies in legal contractual document structure. It also exhibits jurisdictional bias, focusing exclusively on U.S. contracts filed through the SEC. The Red Flag Detection dataset \citep{leivaditi2020benchmark} contains only real estate lease agreements extracted from the U.S. SEC EDGAR system, reflecting both domain and jurisdictional bias and limiting generalizability. Similar biases appear in datasets such as UNFAIR-ToS \citep{lippi2019claudette}, Memnet-ToS \citep{ruggeri2022detecting}, LEXDEMOD \citep{sancheti2022agent}, Contract Ambiguity \citep{singhal2024generating}, Norm \citep{aires2021norm}, ContractNLI \citep{koreeda2021contractnli}, and CUAD \citep{hendrycks2021cuad}, which rely heavily on public U.S. or EU legal contractual documents and focus on narrow contract types such as NDAs, lease agreements, or online terms of service. These observations highlight the challenge of building datasets that span multiple jurisdictions and diverse contract types, especially when working with legal contractual documents that are often restricted by privacy concerns. Addressing annotation noise and domain and jurisdictional bias in future work can enable the development of high-quality, broadly applicable benchmark datasets for the research community.

\emph{\textbf{Pre-processing Legal Contracts:}} Pre-processing legal contracts for classification is a complex task due to the intricate structure and references inherent in legal texts. These documents often contain nested clauses and refer to external legal sources, which pose challenges when trying to break them down into manageable components for analysis. Simply fine-tuning language models on raw legal data is inefficient and impractical, as these contracts require extensive cleaning and transformation to be usable by machine/deep-learning models \citep{ariai2024natural}. Without addressing these structural and contextual complexities, working with large legal contract datasets becomes difficult, limiting the potential for NLP applications in legal fields like contract classification.

\emph{\textbf{Restriction of Multi-Task Learning and Task Diversity}}: A significant challenge in legal contract classification is the absence of a benchmark dataset that supports multi-task learning, where models must perform a variety of tasks simultaneously. Existing datasets mostly focus on one task at a time, which limits model generalization. A more robust dataset should include a range of tasks, such as topic classification, ambiguous clause identification, and deontic modality classification, among others. Furthermore, incorporating fine-grained multi-task classification would enable the evaluation of models on more nuanced aspects, such as distinguishing between different levels of contractual clauses. These diverse tasks are essential for developing models capable of handling the full complexity of real-world legal contracts.

\emph{\textbf{Challenges with Small-size Publicly Available Datasets}}:  
Another challenge lies in the small size of publicly available datasets. For example, Contract Ambiguity \citep{singhal2024generating} contains only 1,000 samples, which is insufficient for method or robust model testing. Although it is acknowledged that legal contract labeling requires expert knowledge—a process that can be costly—training/testing methods or models on such small datasets raises concerns about the reliability and generalizability of the results. A model tested on such a limited corpus may not produce consistent results when applied to larger, more varied contract datasets. Similar issues are present in datasets like Norm \citep{aires2017norm}, which also suffers from a small sample size. To address these limitations, the field would benefit greatly from larger, more diverse datasets that better reflect the complexity and real-world challenges of legal contract classification.

\emph{\textbf{Challenges with Proprietary Datasets}}: Non-public datasets, such as Oblig \& Prohb \citep{chalkidis2018obligation}, and proprietary datasets, such as Contract Requirement \citep{sainani2020extracting} and Fine-grained Obligation \citep{singh2024data}, pose significant challenges due to their lack of public availability for research. We recognize that proprietary datasets are often confidential and restricted to internal company use, as public disclosure may breach contractual agreements. However, these limitations hinder reproducibility, obstruct benchmarking, and restrict the broader research community’s ability to validate, compare, and improve existing models. One potential solution involves organizations using their proprietary datasets to label publicly available contract documents, such as those found in CUAD \citep{hendrycks2021cuad}, which contains 510 contracts. By training models on internal data and applying them to annotate open-access documents, organizations can generate labeled datasets, validate a small sample for performance, and share the results with the wider research community. If direct disclosure of proprietary labels is not possible due to confidentiality concerns, organizations can substitute them with similar or equivalent label types that do not reveal sensitive information. Such initiatives promote transparency, enhance collaboration, and accelerate progress in legal contract classification.
\subsection{Challenges and Future Directions in Terms of Methodology}
\emph{\textbf{Supervised Fine-Tuning in different types of Transformer Architectures:}} As discussed in Section \ref{sec5.3.3}, most studies so far concentrate solely on encoder-based models, with little to no exploration of encoder-decoder models, such as T5 and BART, or decoder-based models, such as LLaMA-3 and Mistral, in the context of contract classification. Future research can address this gap by comparing these different model architectures. In particular, it would be valuable to investigate the performance of encoder-based, encoder-decoder, and decoder-based models across various legal contract classification tasks and identify which type of Transformer architecture proves most effective for handling specific challenges within contract analysis.

\emph{\textbf{Evaluation of Legal Transformer Models:}} Numerous legal LLMs are available today, including Legal-BERT \citep{chalkidis2020legal}, Legal-RoBERTa \citep{geng2021legal}, CaseLawBERT \citep{zheng2021does}, Legal-XLM-Roberta-Large \citep{Niklaus2023MultiLegalPileA6}, ChatLaw \citep{cui2024chatlaw}, AdaptLLM \citep{cheng2023adapting}, PoLBERT \citep{henderson2022pile}, InLegalBERT \citep{paul-2022-pretraining}, LexLM \citep{chalkidis2023lexfiles}, LexT5 \citep{santosh2024lexsumm}, LexGPT \citep{lee2023lexgpt}, InCaseLawBERT \citep{paul-2022-pretraining}, Lawyer-LLaMA \citep{huang2023lawyer}, DISC-LawLLM \citep{yue2023disc}, SaulLM \citep{colombo2024saullm}, CONTRACTS-BERT \citep{chalkidis2020legal}, and others. However, currently no studies broadly assess and compare the performance of these legal LLMs. Future research explores the effectiveness of legal-specific models in contrast to general-purpose models that are not pre-trained on domain-specific legal corpora. The goal is to determine which tasks benefit most from legal-specific models and which tasks can be adequately handled by general-purpose models without requiring legal LLMs. This comparison helps assess whether legal LLMs offer significant advantages across diverse legal contract classification (LCC) tasks and other legal text processing applications.

\emph{\textbf{Strategies for Managing Class Imbalance in LCC:}} Many LCC datasets, such as UNFAIR-ToS \citep{lippi2019claudette}, Red Flag Detection \citep{leivaditi2020benchmark}, and Memnet-ToS \citep{ruggeri2022detecting}, are highly imbalanced, with the majority of instances belonging to non-risky or fair categories. This class imbalance presents a significant challenge for model training, as standard learning algorithms tend to be biased toward the majority class. To address this, various techniques can be applied in future work. For example, data augmentation methods such as DALE \citep{ghosh-etal-2023-dale}, or clustering techniques (e.g., K-Means, DBSCAN, or transformer-based embeddings with agglomerative clustering), can be used to group clauses and identify unlabeled or weakly labeled clauses that resemble existing minority class examples, such as unfair or risky clauses. These clusters can then be labeled using weak supervision or label propagation, thereby expanding the training data without requiring extensive manual annotation \citep{freitas2024text}. The resulting enriched dataset can be used in a supervised learning framework and combined with additional techniques such as class weighting (which assigns higher penalties to misclassified minority class instances), focal loss (which focuses learning on hard-to-classify examples), or balanced sampling to further mitigate class imbalance. Although the current work includes only limited efforts in this direction \citep{ghosh-etal-2023-dale, singhal2023towards}, these techniques represent promising avenues for improving performance in future studies.

\emph{\textbf{Current Challenges in Prompting Strategies and Emerging Research Directions:}}
Although prompt-based methods are increasingly popular for legal contract classification (LCC), they continue to face significant challenges. Large language models (LLMs) such as GPT-3.5 and GPT-4 often make critical errors when processing legal contractual texts. As demonstrated by \cite{wang2024metacognitive}, these models sometimes misinterpret statutes (statutory misinterpretation) or deviate from established legal reasoning (jurisprudential drift). These issues often arise from the inherent complexity of legal language and the nuanced reasoning it requires. This underscores the need for tailored adjustments in metacognitive prompting (MP) to support legal applications more effectively. In particular, prompt engineering techniques \citep{ye2024prompt, marvin2023prompt} must be specifically adapted to legal contexts. \cite{chalkidis2023chatgpt} show that zero-shot and few-shot prompting perform poorly on tasks such as topic classification and unfair clause detection, especially when compared to fine-tuned, domain-specific models. These findings suggest that general-purpose LLMs in zero and few-shot setting, remain not-suited for LCC tasks that demand fixed, nuanced label sets. Similarly, \cite{savelka2023unreasonable} observe that zero-shot prompting mislabels complex legal contractual clauses and fails to capture the subtle semantics necessary for accurate classification. These limitations highlight the inadequacy of current prompting strategies and point to the need for more advanced prompt design or domain-specific pre-training and fine-tuning.

To enhance the prompting-based methods, current research increasingly emphasizes collaboration between legal experts and AI researchers to develop systematic prompt design methodologies. Legal professionals contribute domain-specific insights to create contextually appropriate and precise prompts, while AI researchers apply methodological principles to improve clarity, relevance, and model interpretability. Several recent studies lay the foundation for such approaches. For example, works such as \cite{wang2024prompt, siino2024gpt, chen2023unleashing, velasquez2023prompt} outline principles for prompt construction, iterative testing, error minimization or reducing hallucination, and enhancing reliability and reproducibility. Moreover, existing GPT variants such as GPT-3.5 and GPT-4 are general-purpose. Future research should explore the development or evaluation of legal-specific encoder-decoder and decoder-based LLMs. Assessing how these legal-specific models perform under prompting-based techniques across diverse LCC tasks can help identify which architectures and prompting strategies are most effective for the Contractual NLP domain. This direction promises significantly improved performance and greater alignment with the complex requirements of legal contractual clauses understanding.

\emph{\textbf{Handling Model Limitations and Failure Modes:}} Despite recent advancements in LCC modeling, several challenges remain that limit model effectiveness and generalizability. One key limitation is the difficulty in accurately handling nested or cross-referenced clauses \citep{singh2024data}, which often require tracking dependencies across multiple, noncontiguous parts of a document. Models frequently struggle with these structures due to their limited ability to capture complex discourse relations. One potential solution is the use of hierarchical or graph-based models that represent document structure more explicitly \citep{yu2021knowledge, wang2022d2gclf, paul2022lesicin}. Additionally, integrating coreference resolution and link analysis techniques \citep{lee2018higher} helps models trace relationships between clauses across a document.

Long-range dependencies, where the context necessary for correct interpretation spans several paragraphs or sections, also pose significant challenges for sequence-based architectures, even for transformer-based models \citep{chalkidis2022lexglue}. To address this, researchers use long context transformer architectures such as Longformer or BigBird, as well as hierarchical attention mechanisms \citep{chalkidis2022lexglue}. Alternatively, retrieval augmented methods, which dynamically fetch relevant context during inference \citep{lewis2020retrieval}, also show promise in mitigating the limitations of fixed length input windows.

Finally, legal texts often contain jurisdiction specific terminology, where the same term may carry different legal meanings across regions. This variation significantly impacts model generalization. To manage this, models are trained or fine-tuned on jurisdiction specific data, and external knowledge sources such as legal dictionaries or ontologies are incorporated to provide additional semantic grounding \citep{montemagni2010semantic, palmirani2011legalruleml}. Additionally, developing multilingual or multijurisdictional datasets improves a model’s ability to distinguish and adapt to diverse legal systems.

\emph{\textbf{Ethical Implications and Risks in Automated Legal Contract Classification Systems:}}
The automation of legal contract classification introduces ethical and legal concerns that require careful attention to ensure responsible use. A key risk lies in the misclassification of legally significant or important clauses, which can lead to incorrect interpretations of contractual obligations, rights, or liabilities. For instance, if a termination clause is misclassified or overlooked, it may result in non-compliance, legal disputes, or financial loss for the parties involved. These risks often stem from the nature of the training data used in classification models. Most systems are trained on large datasets of expired or previous contracts, which may reflect outdated legal standards, jurisdiction-specific language, systemic biases, annotation noise, domain-specific bias, or other limitations \citep{edmond2019just, teichman2023biases}. As a result, the system may misidentify or fail to recognize clauses in modern or non-standard contracts, particularly those involving underrepresented parties. This raises concerns about algorithmic bias and the potential to reinforce inequities in legal interpretation.

Automated classification also changes how legal professionals interact with contracts. While these systems speed up the review process, they may lead to over-reliance on LCC system outputs without sufficient legal scrutiny. Legal professionals must remain actively involved in reviewing and validating classifications, especially in high-stakes areas such as mergers and acquisitions, regulatory compliance, or dispute resolution. Legal contract classification systems function best as assistive tools, intended to support, not replace, human legal judgment. As automated LCC classification becomes more widespread in legal practice, it is essential to embed ethical safeguards and ensure transparency in system design. This includes disclosing system limitations, enabling explainable outputs, and incorporating mechanisms for human override. Addressing these concerns is crucial not only for maintaining accuracy and fairness but also for preserving trust in the responsible use of LCC systems in legal contexts.

\emph{\textbf{Balancing Privacy and Performance in Legal AI Applications:}} Privacy plays a major role in the use of AI models for analyzing legal documents, which frequently contain sensitive, confidential, and personally identifiable information such as contracts \citep{solove2025artificial}. A significant barrier to accessing and utilizing such legal data arises from strict data protection laws like the General Data Protection Regulation (GDPR) and the California Consumer Privacy Act (CCPA), which impose legal obligations on entities handling personal information \citep{pazhohan2023global}. These regulations not only govern how data can be collected and processed but also limit the availability of authentic datasets \citep{pazhohan2023global}. To address these concerns, techniques such as differential privacy introduce noise to data during model training to obscure individual records \citep{abadi2016deep}, while adversarial training prepares models to resist inference attacks by suppressing identifiable signals in learned representations \citep{madry2018towards}. In addition, methods like federated learning and privacy-aware fine-tuning enable AI systems to adapt to new tasks without requiring centralized data sharing \citep{mcmahan2017communication, zhuang2023foundation}. However, these solutions require precise tuning, as excessive privacy measures often cause a notable drop in performance, creating a trade-off between data protection and model effectiveness \citep{zhang2016privacy, danezis2015privacy}. This highlights the ongoing need for specialized privacy-preserving strategies, where the complexity and sensitivity of legal language demand nuanced technical and regulatory handling.

\emph{\textbf{Advancement of XAI in Legal Applications:}} Explainable AI (XAI) is crucial for building trust and ensuring the safe deployment of AI technologies, especially in sensitive domains such as LCC. However, research on XAI techniques, particularly in the context of legal applications, remains limited \citep{ariai2024natural, richmond2024explainable}, with even fewer studies focusing on LCC \citep{sen2020learning, ruggeri2022detecting}. Recent work by \cite{sen2020learning} introduces RuleNN, a rule-based, interpretable model that uses logical linguistic patterns for legal contract classification and achieves competitive performance compared to opaque models such as Bi-LSTM. While RuleNN provides clear explanations through rule logic, its reliance on hand-crafted expressions may limit scalability. Augmenting such models with LLM-generated candidate rules (e.g., via Metacognitive Prompting (MP) \citep{wang2024metacognitive}, Chain of Thought (CoT) \citep{wei2022chain}, or Tree of Thoughts (ToT) \citep{yao2024tree}) can improve rule coverage while reducing manual effort. A more promising approach involves combining rule-based methods with transformer-based classifiers and evaluating their predictions using post hoc XAI methods like SHAP \citep{mosca2022shap} and LIME \citep{garreau2020explaining}. Future work explores aligning LLM-generated reasoning steps with feature attribution maps to enhance transparency in legal NLP systems.

\emph{\textbf{Towards Multilingual Legal Contract Classification:}} Most existing research on legal contract classification focuses on English-language contracts. This is mainly because high-quality annotated datasets and pre-trained language models are more readily available for English. As a result, this survey also focuses on English contracts and methods. While this approach allows for a deeper and more focused analysis, it limits the broader applicability of legal NLP systems, especially in multilingual legal environments. In many regions, such as the European Union, legal contracts are written in several official languages. Models trained only on English often do not perform well in these settings. To build more inclusive and adaptable legal NLP systems, it is important to expand research to cover multiple languages. Cross-lingual models like XLM-R \citep{conneau2020unsupervised} and mBERT \citep{devlin2019bert}, as well as recent work by \citet{braun2022clause} on classifying clauses in German contracts, show potential for multilingual contract classification. However, challenges remain. Legal terms, structures, and meanings can vary widely between languages and legal systems. Aligning these differences requires more research to develop models that work well across languages and jurisdictions. Addressing these issues is key to making legal contract classification systems more globally applicable and effective.

\emph{\textbf{ Small Language Models (SLMs) for Legal Contracts Classification:}} There is a notable gap in research on Small Language Models (SLMs) tailored specifically to the Contractual NLP domain \citep{ariai2024natural, wang2024comprehensive}, particularly in the area of legal contract classification. Developing SLMs for this purpose could offer more resource-efficient solutions without sacrificing performance. Such models would enhance the scalability and accessibility of legal NLP tools, making them more affordable and practical for a wider range of users, including smaller law firms and legal tech startups.

\section{Conclusions}\label{sec9}
Research on legal contract classification sees substantial growth in recent years. This paper offers a comprehensive overview of the field, detailing seven distinct tasks, fourteen types of datasets, and thirty-five approaches for automating legal contract classification. These approaches are organized into three main categories: Traditional Machine Learning, Deep Learning, and Large Language Models (LLMs). Among these, multi-class and multi-label classification tasks are the most common. We compile a table summarizing the reported datasets and approaches, providing an organized snapshot of the current state of research. Additionally, we review the evaluation metrics and performance results from various studies, presenting them in a clear and structured manner.

While this research highlights significant progress, it also reveals several key challenges that need to be addressed. These include limitations with existing datasets, the need for higher-quality annotations, and more comprehensive benchmark datasets. Moreover, improving the interpretability and explainability of models remains a critical area for development. Another emerging area of interest is the potential of small language models to enhance legal natural language understanding.

The future of legal contract classification depends on interdisciplinary collaboration to tackle these challenges. Such efforts will lead to the development of more robust, reliable, and scalable systems that can aid in automating the processing and decision-making involved in legal contract documents. By improving the efficiency and accuracy of classification, these systems have the potential to streamline legal workflows, reduce errors, and save time, thereby making legal services more accessible and effective. This could benefit a wide range of users, from commercial enterprises to legal firms and law students, and provides a practical reference for practitioners such as lawyers, compliance experts, contract managers, and legal tech startups seeking to implement or enhance automated legal contract analysis in real-world applications.
\bibliography{sn-bibliography}
\end{document}